\documentclass[lettersize,journal]{IEEEtran}

\usepackage{multirow}
\usepackage{amsmath,amsfonts}
\usepackage{array}
\usepackage{subcaption}
\usepackage[caption=false,font=normalsize,labelfont=sf,textfont=sf]{subfig}
\usepackage{textcomp}
\usepackage{stfloats}
\usepackage{url}
\usepackage{verbatim}
\usepackage{graphicx}
\usepackage{cite}
\usepackage {algpseudocode}
\usepackage[ruled,vlined,linesnumbered]{algorithm2e}
\usepackage{algorithmicx}
\usepackage{algcompatible}
\usepackage{amsmath}
\usepackage{color}
\usepackage{xcolor}

\newcommand{\tB}[1]{\textcolor{blue}{\textbf{#1}}}

\newcommand{\Cgray}[1]{\cellcolor[HTML]{D8D6D6}} %

\begin{document}

\title{Object Re-identification via Spatial-temporal \\ Fusion Networks and Causal Identity Matching}

\author{Hye-Geun~Kim,~\IEEEmembership{Student Member,~IEEE}, Yong-Hyuk~Moon,
	and~Yeong-Jun~Cho,~\IEEEmembership{Member,~IEEE}
	\IEEEcompsocitemizethanks{\IEEEcompsocthanksitem Hye-Geun Kim is with the Department of AI convergence, Chonnam National University, South Korea.
		Yong-Hyuk Moon is with Electronics and Telecommunications Research Institute, Daejeon, South Korea.
		Yeong-Jun Cho is with the Department of AI convergence, Chonnam National University, South Korea. (e-mail: yj.cho@chonnam.ac.kr). \protect
	}
	}

\markboth{Journal of \LaTeX\ Class Files,~Vol.~0, No.~0, August~2024}%
{Shell \MakeLowercase{\textit{et al.}}: A Sample Article Using IEEEtran.cls for IEEE Journals}

\IEEEpubid{0000--0000/00\$00.00~\copyright~2024 IEEE}
\IEEEpubidadjcol

\maketitle

\begin{abstract}
Object re-identification~(ReID) in large camera networks faces numerous challenges.
First, the similar appearances of objects degrade ReID performance, a challenge that needs to be addressed by existing appearance-based ReID methods.
Second, most ReID studies are performed in laboratory settings and do not consider real-world scenarios. 
To overcome these challenges, we introduce a novel ReID framework that leverages a spatial-temporal fusion network and causal identity matching (CIM).
Our framework estimates camera network topology using a proposed adaptive Parzen window and combines appearance features with spatial-temporal cues within the fusion network.
This approach has demonstrated outstanding performance across several datasets, including \texttt{VeRi776}, \texttt{Vehicle-3I}, and \texttt{Market-1501}, achieving up to 99.70\% rank-1 accuracy and 95.5\% mAP.
Furthermore, the proposed CIM approach, which dynamically assigns gallery sets based on camera network topology, has further improved ReID accuracy and robustness in real-world settings, evidenced by a 94.95\% mAP and a 95.19\% F1 score on the \texttt{Vehicle-3I} dataset. 
The experimental results support the effectiveness of incorporating spatial-temporal information and CIM for real-world ReID scenarios, regardless of the data domain (e.g., vehicle, person).
\end{abstract}


\section{Introduction}
\label{sec:intro}

\IEEEPARstart{A}{n} increasing surveillance cameras have been installed in public places for safety, traffic monitoring, and security. 
However, monitoring all cameras requires significant human effort and resources.
Re-identification~(ReID) can automatically track targets across multiple non-overlapping cameras, reducing human effort.
In general, most studies have focused on the visual appearance of a target to perform ReID.
For example, many studies~\cite{ahmed2015improved, chen2017person} have proposed feature learning methods to represent the appearance of a target. 
Similarly, to measure feature distance between queries and gallery images effectively, metric learning methods~\cite{zheng2011person, koestinger2012large} have also been proposed. 
Recently, ReID performance has been improved by learning visual features and distance metrics with developments in deep learning~\cite{he2020fastreid,ghosh2023relation}. 
These appearance-based methods are robust to pose variations, viewpoint changes, and illumination changes.

\begin{figure}
  \centering
  \begin{subfigure}{1\linewidth}
    \includegraphics[width=\linewidth]{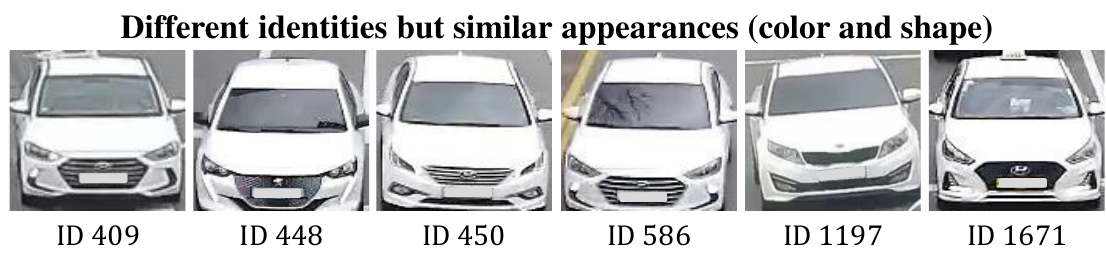}
    \caption{Appearance ambiguities}
  \end{subfigure}
  \hfill
  \begin{subfigure}{1\linewidth}
    \includegraphics[width=\linewidth]{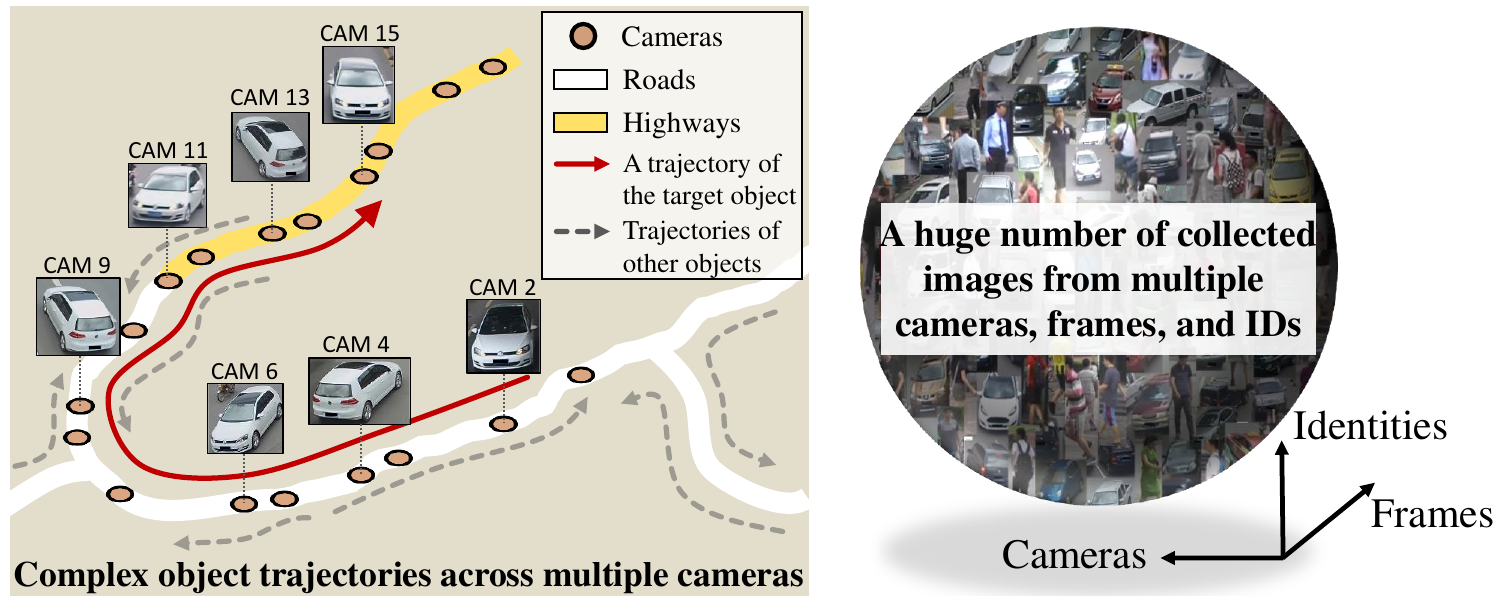}
    \caption{High computational complexities}
  \end{subfigure}
  \caption{Challenges of object re-identification in large-scale camera networks due to appearance ambiguities and computational complexities}
  \label{fig:1}
\end{figure}

Nevertheless, appearance ambiguity caused by objects with similar appearances is still not alleviated. 
As shown in Fig.~\ref{fig:1}a, vehicles are challenging to match accurately due to their similar appearances, especially when they are of the same model type. 
In contrast, people generally have more distinct features, but similar appearances can still arise when they wear the same clothing.
Additionally, the appearance of many objects across multiple cameras increases computational complexity and reduces identification performance as the number of objects with appearances similar to the target increases (Fig.~\ref{fig:1}b). 
Thus, relying only on target appearance is ineffective for the object ReID problem.

\IEEEpubidadjcol

Recently, ReID studies using additional spatial and temporal information to alleviate appearance ambiguity~\cite{cho2019joint, shen2017learning, huang2022vehicle, wang2019spatial} have been proposed.
Researchers have built a camera-network topology that explains the spatial and temporal relationships between cameras and used the topology to reduce redundant retrieval time ranges for queries. 
While these methods can potentially improve appearance-based ReID models~\cite{he2020fastreid,ghosh2023relation}, they still have limitations. 
Their approaches to camera network topology modeling are too simple, and integrating the appearance model with spatial-temporal information lacks optimization.
In addition, most ReID studies have not considered applying ReID in real-world scenarios, such as handling real-time video continuously coming from each camera. 
However, most ReID studies~\cite{he2020fastreid, he2021transreid} have performed ReID on fixed and refined query-gallery sets, which are clean and easy to manage.
These laboratory settings are difficult to implement in real-world environments and require additional processes such as object appearance management and query-gallery sets management.
In summary, previous studies have not fully considered the spatial-temporal dependencies between cameras and the complexities of real-world scenarios.

To consider the spatial and temporal dependencies between cameras, we proposed a spatial-temporal fusion network (FusionNet) in our previous study~\cite{kim2023spatial}.
For training FusionNet, we proposed a novel adaptive Parzen window method robust to outliers and sparse responses between camera pairs (Section~\ref{sec:proposed_1}). 
This method effectively manages varying connection strengths between camera pairs, enabling reliable camera network topology estimation.
After estimating the topology, we trained FusionNet to optimally combine appearance similarity and spatial-temporal probabilities (Section~\ref{sec:proposed_2}).
In this study, we extend our study\cite{kim2023spatial} to explore ReID problems in real-world scenarios extensively. 
To this end, we first built a new large-scale vehicle ReID dataset called \texttt{Vehicle-3I}, which includes 2,038 identities captured by 11 synchronized cameras installed at three different intersections (Section~\ref{sec:datasets}). Unlike many public datasets~\cite{zheng2015scalable, liu2016deep} that provide only cropped image patches, it provides full-frame videos. 

Furthermore, we focus on critical and practical issues for ReID in real-world scenarios.
Issue 1): Each identity appears multiple times in the video, necessitating an appearance management process and ID-to-ID ReID.
Issue 2): The query and gallery sets cannot be predefined as in previous studies. 
Instead, they are determined dynamically based on the target objects~(query) and their ReID results within the camera network.
In Section~\ref{sec:real-world}, we propose appearance management and ID-to-ID ReID methods that consider multiple appearances of each identity in the video. 
Additionally, we implement Causal Identity Matching (CIM), which dynamically adjusts the query and gallery sets according to the camera connection causality.

To evaluate the proposed spatial-temporal fusion network, we tested it on the \texttt{VeRi776}~\cite{liu2016deep}, \texttt{Vehicle-3I} vehicle ReID datasets and the \texttt{Market-1501}~\cite{zheng2015scalable} person ReID dataset.
We evaluated the effectiveness of the adaptive Parzen window and FusionNet. 
The proposed spatial-temporal fusion network achieved the best performances across all datasets for rank-1 accuracy (99.70\%, 89.22\%, 99.11\%) and mean average precision~(mAP) (91.71\%, 83.62\%, 93.80\%).
These results demonstrate that our methods can significantly improve ReID performance across different data domains (e.g., vehicle, person).
Additionally, we performed CIM and evaluated the proposed methods in real-world scenarios; the proposed methods achieved superior performances on the \texttt{Vehicle-3I} dataset, with a 94.95\% mAP and 95.19\% $F_1$ score.

The contributions of our study are as follows:
\begin{itemize}
	\item Proposing FusionNet that combines two different similarities (appearance and spatial-temporal).
	\item Creating a new large-scale vehicle ReID dataset.
	\item Considering real-world ReID scenarios and performing extensive evaluations.
	\item Achieving superior performances on both vehicle and person ReID tasks.	
\end{itemize}

\section{Related Works}
\label{sec:related_work}

\subsection{Appearance-based ReID}

Most ReID studies have focused on developing visual representations of images to distinguish their appearance. 
Extensive research has been conducted on metric learning and feature learning methods to achieve this objective. 
A widely studied approach in metric learning involves learning the Mahalanobis distance~\cite{zheng2011person, koestinger2012large}.  
Specifically, optimizing triplet loss for deep metric learning~\cite{cheng2016person,hermans2017defense, he2020fastreid} has demonstrated outstanding performance in ReID tasks. 
Ghosh et al.~\cite{ghosh2023relation} introduced relation-preserving triplet mining (RPTM), a scheme for feature matching-guided triplet mining that ensures triplets preserve natural subgroupings in object IDs.

Ahmed et al.~\cite{ahmed2015improved} initially employed a deep convolutional neural network~(CNN) architecture to capture local relationships between two input images based on mid-level features.
Chen et al.~\cite{chen2017person} improved CNN-based ReID by proposing a deep pyramid feature learning CNN, which can learn scale-specific discriminative features.
Similarly, other studies~\cite{cheng2016person,rong2021vehicle} have focused on extracting robust local features.
Khamis et al.~\cite{khamis2015joint} utilized target attributes for ReID to capture more appearance details.
Recently, He et al.~\cite{he2021transreid} proposed TransReID, marking the first attempt to use a transformer to learn robust features from images. 
Chen et al.~\cite{chen2023beyond} utilized the Swin transformer~\cite{liu2021swin} as a backbone for downstream ReID tasks.

Li et al.~\cite{li2023clip} proposed CLIP-ReID, which fine-tunes the initialized visual model using the image encoder in CLIP for improved visual representation. 
Additionally, other studies~\cite{yang2022unleashing,zhu2022pass} have focused on pre-training methods to overcome the domain gap in ReID tasks. 
Furthermore, several approaches~\cite{cho2016improving, su2017pose, miao2019pose, he2020guided} have utilized extra cues, such as human pose and body parts, to address pose variations and occlusion issues. 
Similarly, studies~\cite{zhu2020identity,huynh2021strong} have aimed to improve visual representation quality through additional identity-guided human semantic parsing and multi-head attention.
Nevertheless, alleviating appearance ambiguity is still challenging when relying only on appearance for ReID.

\begin{figure*}
	\centering
	\begin{subfigure}{0.45\linewidth}
		\includegraphics[width=\linewidth]{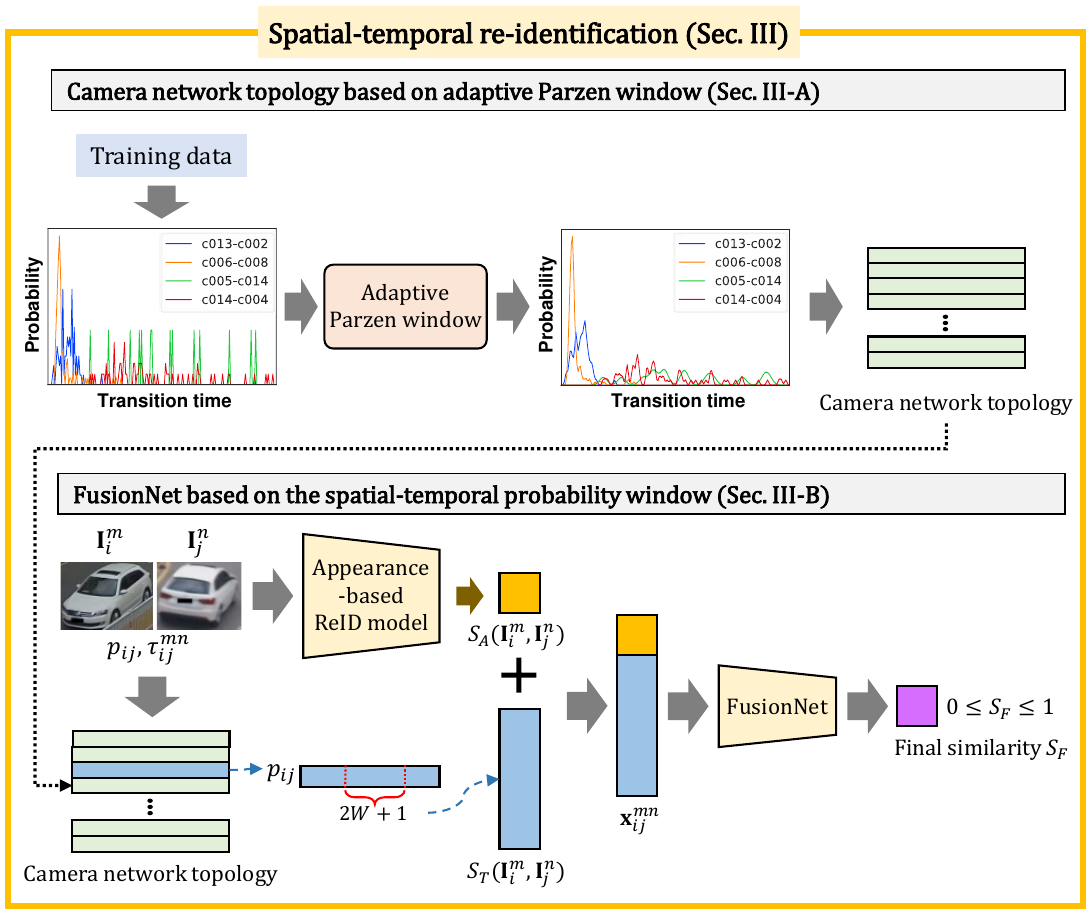}
		\caption{}
		\label{fig:2a}
	\end{subfigure}
	\begin{subfigure}{0.45\linewidth}
		\includegraphics[width=\linewidth]{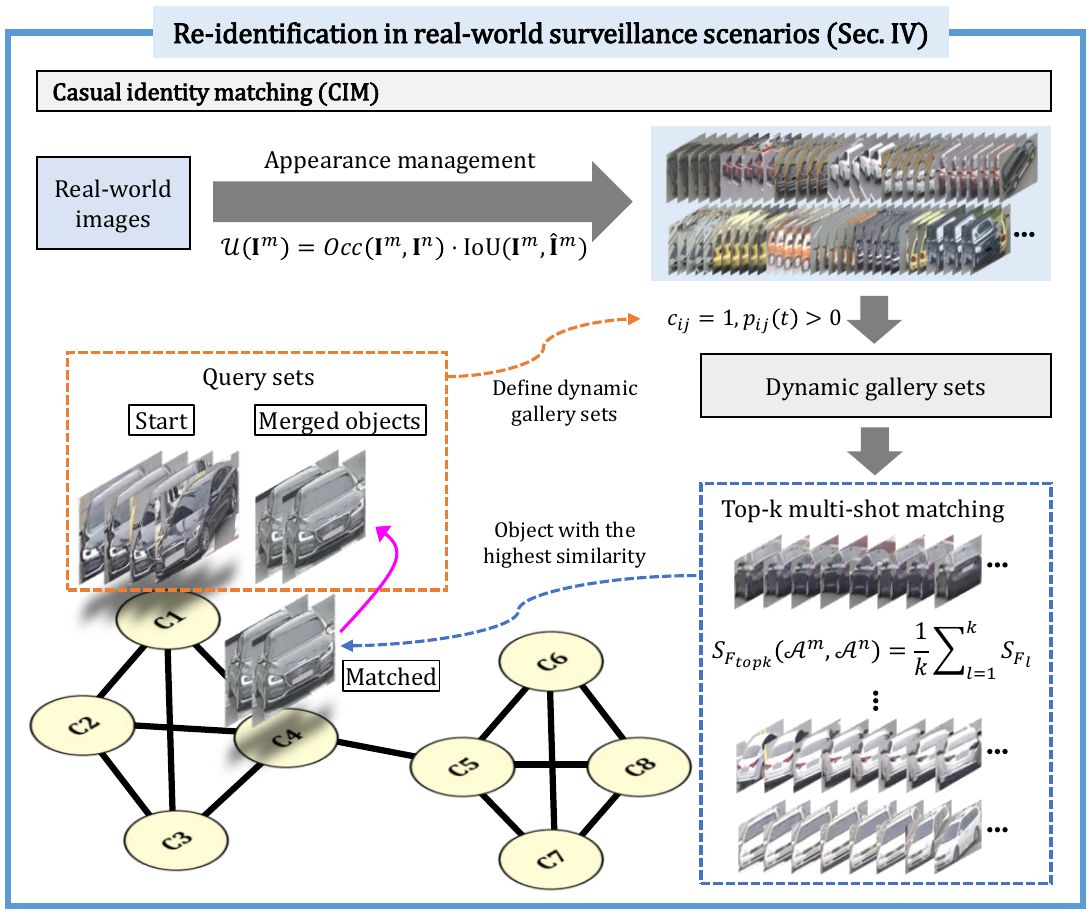}
		\caption{}
		\label{fig:2b}
	\end{subfigure}
	\caption{The overall ReID methods for object re-identification with spatial-temporal information in real-world scenarios.}
	\label{fig:2}
\end{figure*}

\subsection{Spatial-temporal ReID}

Many studies have utilized spatial-temporal information from cameras and target objects to address the limitations of appearance-based ReID.
Generally, these studies take an appearance-based ReID model as the baseline and leverage spatial and temporal information. 
In spatial-temporal ReID, two main challenges arise: 1) estimating spatial-temporal information (the camera network topology) in given camera networks, and 2) utilizing the estimated camera network topology for ReID.

Many studies have attempted to estimate the camera network topology by designing accurate transition time distributions of targets, such as persons and vehicles. 
For instance, Huang et al.~\cite{huang2022vehicle} utilized a spatial-temporal model leveraging vehicle pose view embedding, and Wang et al.~\cite{wang2019spatial} proposed the Histogram-Parzen method to estimate spatial-temporal probability distributions. 
Additionally, Liu et al.~\cite{liu2016deep,liu2017provid} proposed a progressive vehicle ReID that applies simple spatial-temporal information. 
Similarly, studies~\cite{cho2019joint,lv2019vehicle,zheng2020vehiclenet, ren2021learning} have utilized estimated spatial-temporal information to filter out irrelevant gallery images. 
Furthermore, Shen et al.~\cite{shen2017learning} proposed a Siamese-CNN+Path-LSTM network that predicts paths using both visual and spatial-temporal information.  

While numerous spatial-temporal ReID methods have been proposed, limitations remain. 
First, the methodologies for estimating spatial-temporal models are often simplistic. For example, many methods~\cite{cho2019joint,liu2016deep,liu2017provid,lv2019vehicle} have built object transition time distributions based on positive responses between cameras, but noisy and sparse responses make the estimated distributions unreliable. 
Second, the utilization of spatial-temporal information is not optimized.
Some studies~\cite{wang2019spatial,ren2021learning,huang2022vehicle} have merged both probabilities~(i.e., appearance and spatial-temporal) with same importance to obtain the joint probability. 
Similarly, many methods~\cite{cho2019joint,lv2019vehicle,zheng2020vehiclenet, ren2021learning} have applied spatial-temporal information to reduce the search range or perform re-ranking of the initial ReID results.

\section{Spatial-temporal Re-identification \\ based on FusionNet}

To address the challenges in ReID, we analyzed the characteristics of objects in camera networks.
First, many objects show similar or the same appearances in large-scale camera networks.
For example, people may appear similar because they wear the same clothes~(e.g., uniforms and belongings).
Especially in the vehicle ReID task, vehicles can look the same according to their model types.
Second, the movements of objects between non-overlapping cameras are predictable. 
For instance, vehicles can only move along roads and highways, rarely deviating from existing routes.
Compared to vehicles, people show more complex transition patterns across cameras, but people's paths are also generally established along the sidewalks and aisles.
In summary, objects show high appearance ambiguities but predictable movements.
Therefore, relying only on appearance differences between objects is not effective for the ReID tasks.

Based on these observations, we further exploited spatial and temporal relationships between cameras, referred to as camera network topology. 
As shown in Fig.~\ref{fig:2a}, the proposed spatial-temporal ReID framework consists of two parts: 
1) camera network topology estimation based on the adaptive Parzen window, 2) the fusion network on the spatial-temporal probability window.
We first built the topology based on the proposed adaptive Parzen window~(Section~\ref{sec:proposed_1}). 
Then, we trained FusionNet to optimally combine visual similarity with camera network topology information for final ReID prediction~(Section~\ref{sec:proposed_2}).

\begin{figure*}
	\centering
	\begin{subfigure}{0.32\linewidth}
		\includegraphics[width=\linewidth]{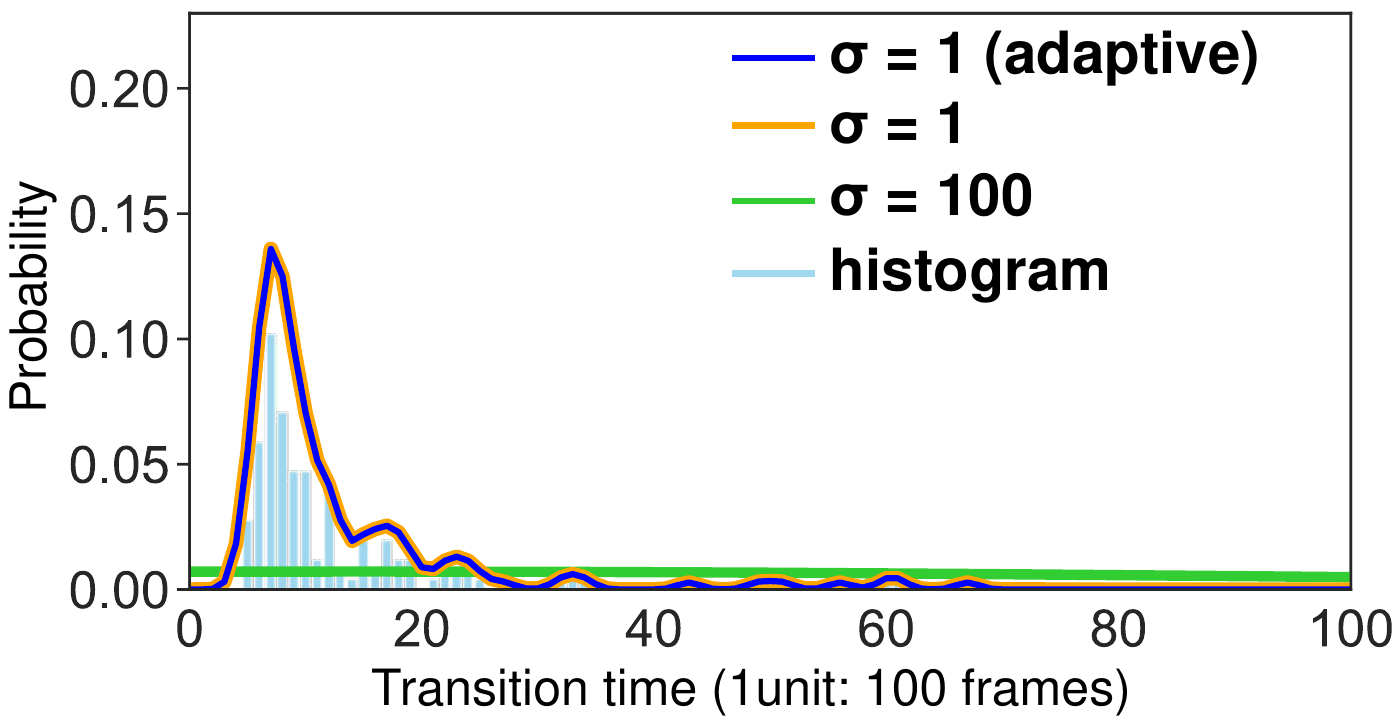}
		\caption{Pairs between cameras: 143\\ \centering (Connection strength: strong)}
		\label{subfig:sub1}
	\end{subfigure}
	\hfill
	\begin{subfigure}{0.32\linewidth}
		\includegraphics[width=\linewidth]{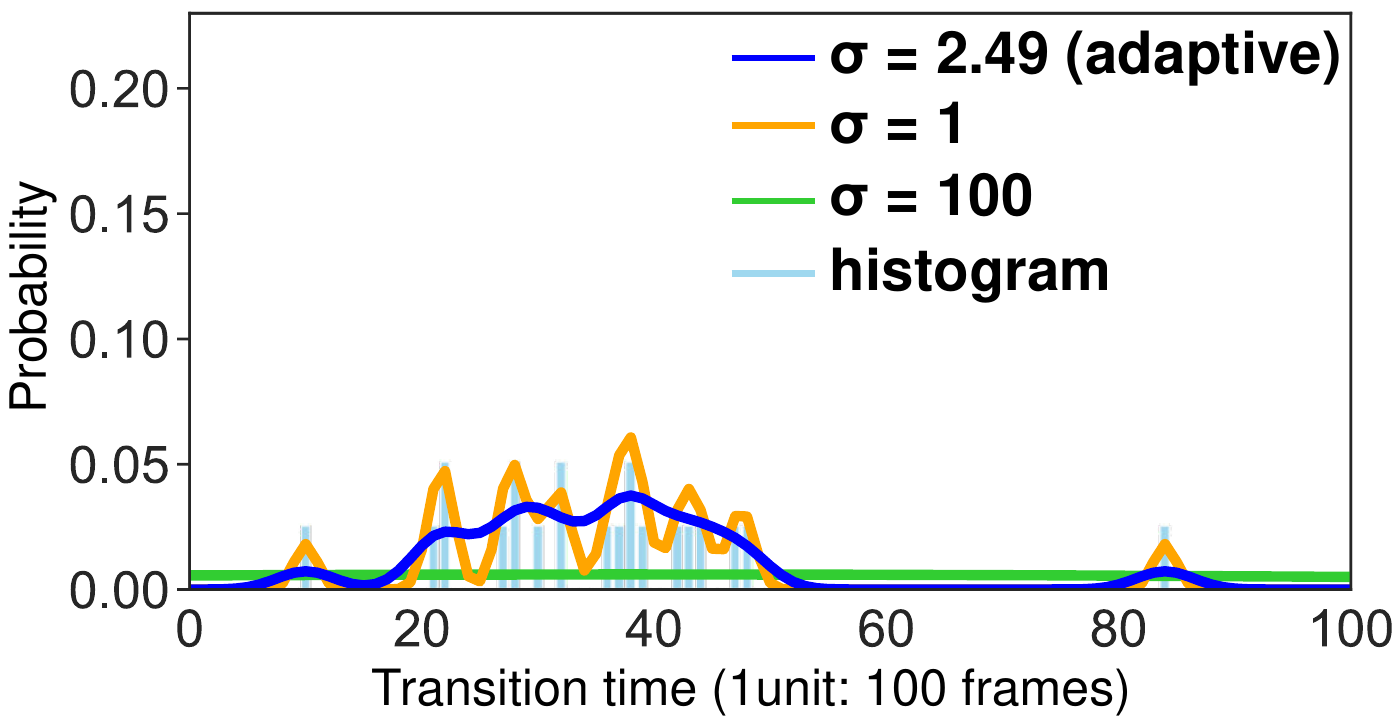}
		\caption{Pairs between cameras: 22\\ \centering (Connection strength: normal)}
		\label{subfig:sub2}
	\end{subfigure}
	\hfill
	\begin{subfigure}{0.32\linewidth}
		\includegraphics[width=\linewidth]{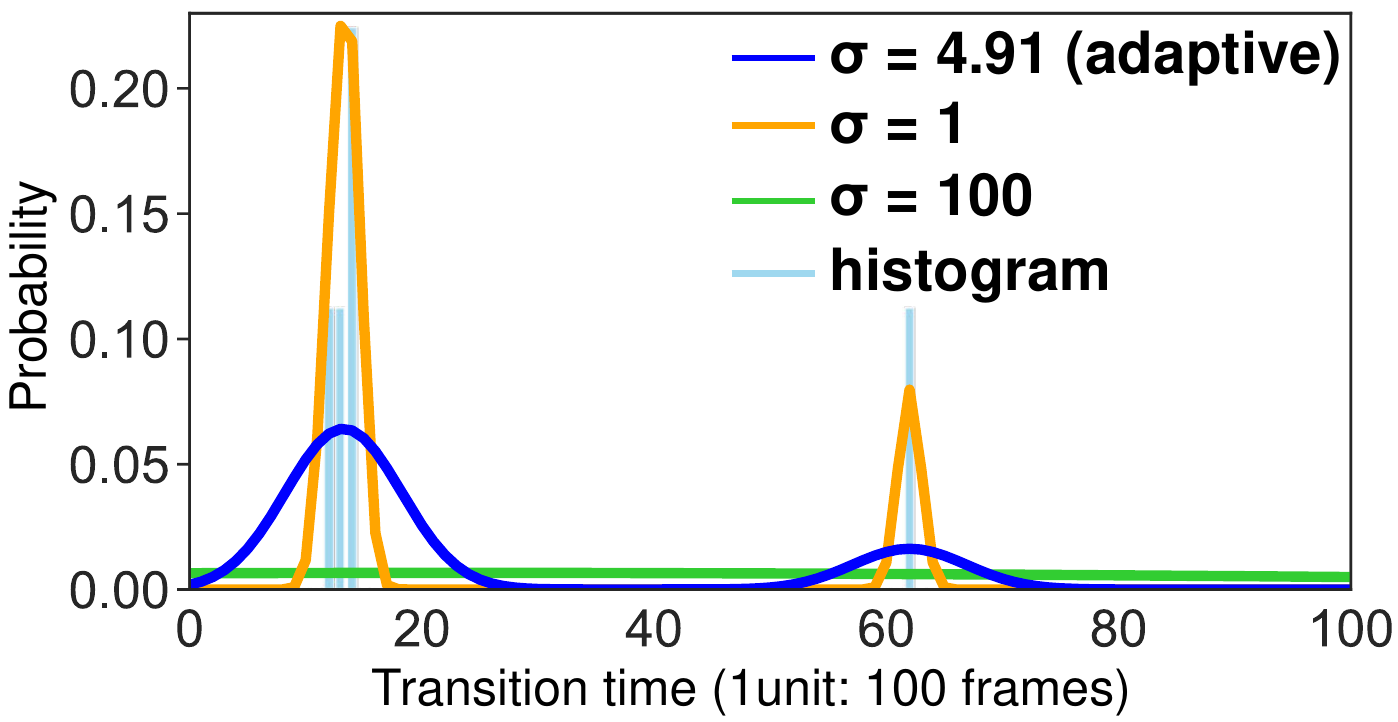}
		\caption{Pairs between cameras: 5\\ \centering (Connection strength: weak)}
		\label{subfig:sub3}
	\end{subfigure}
	\caption{Examples of estimated transition time distributions between camera pairs. Each bin covers 100 frame ranges. Solid blue lines (\tB{---}) mark the estimated distribution~($p_{ij}$) from the histogram~($h_{ij}$) using the proposed adaptive Parzen window (best viewed in color).}
	\label{fig:three_figures}
\end{figure*}

\subsection{Adaptive Parzen window}
\label{sec:proposed_1}

The camera network topology represents the spatial-temporal relationships and connections between cameras, which can be represented as a graph $\mathbf{G} = (\mathbf{V}, \mathbf{E})$. 
The vertices $\mathbf{V}$ represent the cameras, and the edges $\mathbf{E}$ represent the distribution of object transition times. 
If there are $N_{\text{cam}}$ cameras in the network, the topology is represented as follows:    

\begin{equation}
	\begin{split}
		\mathbf{V} & \in \{c_i|1\leq i\leq N_{cam}\}, \\
		\mathbf{E} & \in \{p_{ij}|1\leq i\leq N_{cam}, 1\leq j\leq N_{cam}, i\neq j\ \},
	\end{split} 
	\label{eq:1}
\end{equation}
where $c$ denotes a camera, and $p_{ij}$ denotes the object transition time distribution between the camera pair $c_i$ and $c_j$.

We build the transition time distribution $p_{ij}$ by leveraging the training dataset's positive pairs from all camera pairs. 
Using the multiple time differences~$(\Delta t)$ of positive pairs, we initially generated a histogram of the transition time $h_{ij}$, which is illustrated by the cyan vertical lines (\textbf{\textcolor{cyan}{---}}) in Fig.~\ref{fig:three_figures}. 
Cho et al.~\cite{cho2019joint} proposed connectivity checking criteria for determining if a pair of cameras are connected by fitting a Gaussian model to the histogram $h_{ij}$. 
However, this parametric method relied on strong assumptions and struggled to handle outliers and the sparsity of the histogram. 
Inspired by~\cite{wang2019spatial}, we applied a Parzen window method to the initial histograms and estimated the probability density function~(PDF) of the transition times in a non-parametric manner as follows: 
\begin{equation}
	p_{ij}\left(\tau\right) = \frac{1}{Z} \sum_{l} h_{ij}\left(\tau\right)K\left(l-\tau\right),
	\label{eq:2}
\end{equation}
where $\tau$ is an index of the distribution, $Z=\sum_{l} p_{ij}(\tau)$ represents a normalized factor, and $K(\cdot)$ is a kernel function. 
For the kernel $K$, we used the Gaussian function, as follows:
\begin{equation}
	K(x) = \frac{1}{\sqrt{2\pi}\sigma}\exp\left({\frac{-x^2}{2\sigma^2}}\right),
	\label{eq:3}
\end{equation}
where $\sigma$ is the standard deviation.

Estimating continuous PDFs from discrete histograms efficiently using the Parzen window method is unreasonable when employing a single kernel across diverse histograms from various camera pairs. 
The strength of the spatial-temporal connection between cameras can be determined by the number of objects passing through those cameras during a certain period~\cite{cho2019joint}. 
For example, weak connectivity is indicated by few positive pairs between two cameras. 
However, the Parzen window method can excessively amplify small responses when using a small $\sigma$ value, as depicted by the orange line (\textbf{\textcolor{orange}{---}}) in Fig.~\ref{fig:three_figures}c. 
In that case, it is better to use a large $\sigma$ value to avoid overfitting the distribution to noise and outliers.

In contrast, if there are many positive pairs between cameras, the connectivity should be strong. 
However, when using a large $\sigma$ value, the resulting distribution becomes uniform, failing to capture meaningful spatial and temporal relationships between the cameras, as illustrated by the green line (\textbf{\textcolor{green}{---}}) in Fig.~\ref{fig:three_figures}a. 
In such cases, it is better to use a relatively small $\sigma$ value to reflect temporal information between cameras.
Therefore, selecting the appropriate $\sigma$ value is important for the quality of the estimated distribution $(p_{ij})$.

To overcome the limitations of the original Parzen window method~\cite{wang2019spatial}, we newly propose an adaptive Parzen window by setting various $\sigma_{ij}$ values for the camera pairs~$(c_i,c_j)$. 
To this end, we designed an adaptive standard deviation according to the different strengths of camera connectivity as follows:
\begin{equation}
\sigma_{ij} = \max \left(\alpha \exp\left({\frac{-N_{ij}}{\beta}}\right), 1\right),
\label{eq:4}
\end{equation}
where $N_{ij}$ denotes the number of positive object pairs between cameras $c_i$ and $c_j$. 
Additionally, $\alpha$ is a scale factor determining the maximum range of $\sigma_{ij}$, and $\beta$ is a smoothness factor adjusting the sensitivity of $\sigma_{ij}$. 
The minimum value of $\sigma_{ij}$ cannot be less than 1 unit of the histogram. 
Therefore, the values of $\sigma_{ij}$ lie within the range of $\left[1, \alpha\right]$.

Considering the camera indices, Eq.~\ref{eq:2} and Eq.~\ref{eq:3} are reformulated as follows:

\begin{equation}
	p_{ij}\left(\tau\right) = \frac{1}{Z} \sum_{l} h_{ij}\left(\tau\right)K_{ij}\left(l-\tau\right),
\end{equation}
\begin{equation}
	K_{ij}(x) = \frac{1}{\sqrt{2\pi}\sigma_{ij}}\exp\left({\frac{-x^2}{2\sigma_{ij}^2}}\right).
\end{equation}
Therefore, we can estimate reliable distributions~($p_{ij}$) from the initial discrete histograms~($h_{ij}$) by considering the connectivity between cameras. 
The blue lines (\tB{---}) in Fig.~\ref{fig:three_figures} represent our results based on the adaptive Parzen window.

\subsection{Fusion Network}
\label{sec:proposed_2}

The baseline for the proposed ReID framework can be any appearance-based ReID method that estimates appearance similarities between images.
Object images from each camera pair $(c_i,c_j)$ are represented as $\mathbf{I}^{m}_i$ and $\mathbf{I}^{n}_j$, where $m$ and $n$ are the image indices.
The appearance-based ReID methods then estimate the visual similarity between two images as $S_A(\mathbf{I}^{m}_i,\mathbf{I}^{n}_j)$, which lies within the range $\left[0, 1\right]$. 
The proposed framework does not rely only on typical appearance-based models.

To perform spatial-temporal ReID, Cho et al.~\cite{cho2019joint} used only camera network topology to limit the gallery search range, effectively reducing the complexity of ReID. 
However, the spatial-temporal probability did not affect the final similarity. In contrast, previous studies~\cite{wang2019spatial,ren2021learning,huang2022vehicle} have merged both probabilities (i.e., appearance and spatial-temporal) with the same importance to obtain the joint probability. 
However, they neglected two key points. 
First, the domains of each probability are not the same. 
Second, both appearance and spatial-temporal probabilities can be imperfect. 
Hence, simply merging these probabilities is unreasonable.
	
In this study, we optimally combined visual similarities $S_A(\mathbf{I}^{m}_i,\mathbf{I}^{n}_j)$ and estimated spatial-temporal probability distributions $p_{ij}\left(\tau\right)$ through a fusion network named FusionNet. 
The input vector of the network for two images $(\mathbf{I}^{m}_i,\mathbf{I}^{n}_j)$ in the camera pair $(i,j)$ can be represented by

\begin{equation} 
    \mathbf{x}^{mn}_{ij} = [S_A(\mathbf{I}^{m}_i,\mathbf{I}^{n}_j), S_T(\mathbf{I}^{m}_i,\mathbf{I}^{n}_j)],
    \label{eq:6}
\end{equation}
where $S_A$ is the appearance similarity, and $S_T$ is the spatial-temporal vector.
The $S_T$ vector between the images is defined by
\begin{equation}
    {\fontsize{8.5}{12}
    \begin{split}
        S_T(\mathbf{I}^{m}_i,\mathbf{I}^{n}_j) = \biggl[ & p_{ij}\left(\tau^{mn}_{ij}-W\right), ... , \\
        & p_{ij}\left(\tau^{mn}_{ij}-1\right),p_{ij}\left(\tau^{mn}_{ij}\right),p_{ij}\left(\tau^{mn}_{ij}+1\right), \\
        & ...,p_{ij}\left(\tau^{mn}_{ij}+W\right) \biggr],
    \end{split}}
    \label{eq:7}
\end{equation}
the notation $\tau^{mn}_{ij}$ represents the time difference between two images $\mathbf{I}^{m}_i$ and $\mathbf{I}^{n}_j$.
The parameter $W$ denotes the size of a time window. 
The range of the $S_T$ vector depends on the value of $W$ and is determined around the distributions of $p_{ij}(\tau^{mn}_{ij})$. 
For example, when $W=0$, $S_T$ becomes a scalar value given by $S_T(\mathbf{I}^{m}_i,\mathbf{I}^{n}_j) = p_{ij}(\tau^{mn}_{ij})$. 
If $W>0$, the $S_T$ vector has a dimension of $2W+1$. 
The value of $W$ can be adjusted to determine the amount of spatial-temporal information provided to FusionNet. 
Consequently, the dimension of the input vector $\mathbf{x}^{mn}_{ij}$ for FusionNet is $2W+2$.

\begin{figure*}
   \centering
   \begin{subfigure}{0.4\linewidth}
      \includegraphics[width=\linewidth]{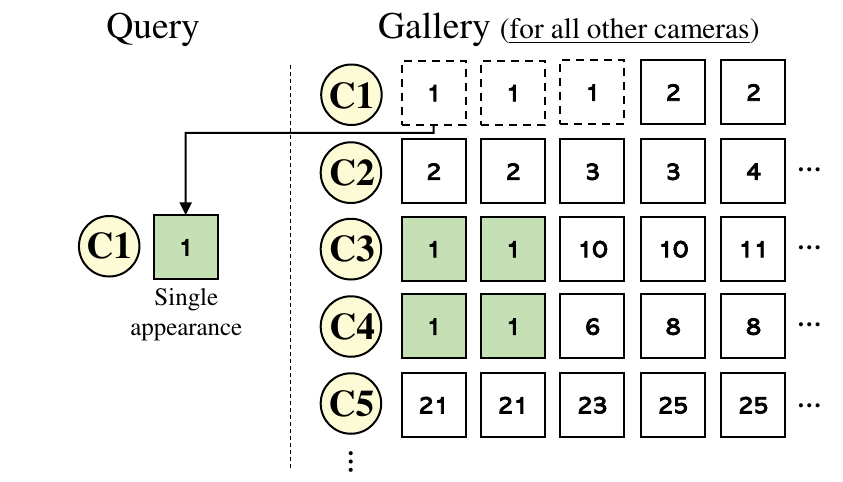}
      \caption{One-to-all ReID without causality}
   \end{subfigure}
   \hspace{30pt}
   \begin{subfigure}{0.4\linewidth}
      \includegraphics[width=\linewidth]{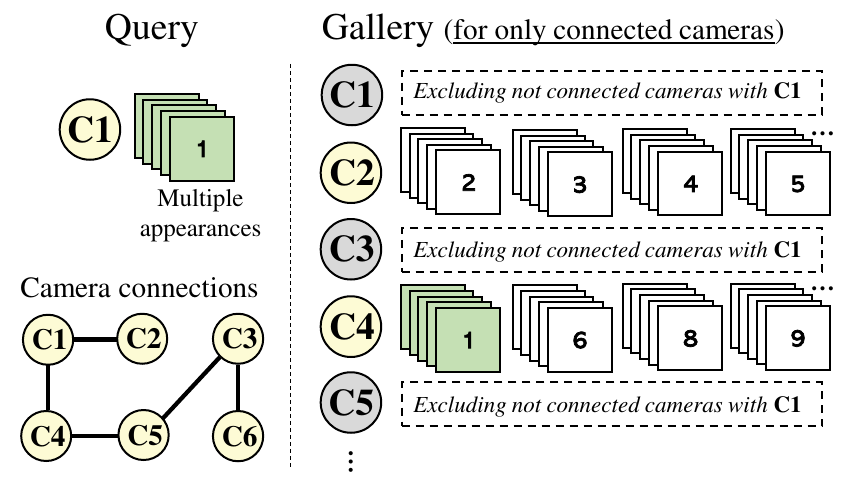}
      \caption{ID-to-ID ReID considering causality}
   \end{subfigure}
   \caption{Comparisons of ReID methodologies: Circles and boxes denote cameras and appearances. The number inside each box represents the appearance ID. Boxes filled with olive color indicate true positives concerning the query.
   Boxes with dotted lines indicate appearances excluded from the gallery.
   (a) One-to-all ReID using only a single appearance for the query performs many redundant and duplicated comparisons, even comparing different objects within the same camera. \\
   (b) ID-to-ID ReID uses multiple appearances and considers causality to determine the gallery.}
   \label{fig:single_multi_matching}
\end{figure*}

The FusionNet was designed based on a simple multi-layer perception. 
Empirical findings revealed that FusionNet does not require a sophisticated deep neural network structure to estimate the final similarity. 
The network consists of one hidden layer with several nodes and a one-dimensional output layer, as illustrated in Fig.~\ref{fig:2a}. 
For the activation function, we used the Rectified Linear Unit (ReLU) for nodes in the hidden layer and the sigmoid function for the output node. 
Consequently, the final output of FusionNet, $S_F(\mathbf{I}^{m}_i, \mathbf{I}^{n}_j)$, is within the range $\left[0,1\right]$. 
To train the network, we optimized the binary cross-entropy loss, defined as follows:

\begin{equation}
    \mathcal{L} =\sum_k{y_k \log{S_F\left(k\right)}+\left(1-y_k\right)\log(1-{S_F\left(k\right))}},	
\end{equation}
where $k$ represents the index of the training image pair, and $y_k\in[0,1]$ denotes the ground truth of the $k$-th image pair.

\section{Re-identification in \\ Real-world Surveillance Scenarios}
\label{sec:real-world}
Previous studies have evaluated their methods using a fixed and refined gallery set.
They selected a query image from the gallery and compared it with all other images in the gallery, which is a one-to-all single-shot comparison.
This evaluation process is repeated until all queries are compared, and then all the results are averaged to compute the final ReID performance.
Note that the gallery set contains multiple appearances for a single identity, leading to many duplicated comparisons as shown in Fig.~\ref{fig:single_multi_matching}a.
These laboratory evaluation settings are unsuitable for real-world scenarios, where multiple appearances of the object do not need to be evaluated individually.
	
Additional issues beyond laboratory settings must be considered in real-world surveillance scenarios. 
First, each identity can have multiple appearances in the video, and these appearances should be evaluated in an ID-to-ID manner, as explained in Fig.~\ref{fig:single_multi_matching}b.
Second, the query and gallery sets are not fixed throughout all test processes; instead, they can change according to the target objects and matching causality.
We propose CIM based on our previous work~\cite{kim2023spatial} to address these challenges in real-world scenarios.
CIM includes three different methods (Sections~\ref{sec:proposed_3}--~\ref{sec:proposed_5}) for effective ReID in real-world scenarios.
The overall framework of CIM is illustrated in Fig.~\ref{fig:2b}.

\subsection{Appearance management for each object}
\label{sec:proposed_3}

As mentioned above, each object has multiple appearances along its trajectory in the sequence.
Assuming that a camera captures multiple frames per second, we can collect various images, which is advantageous for ReID. 
However, this also requires a significant amount of computation. We observed that many of the appearances are unreliable and redundant in many cases.
For example, when one object occludes another, the appearances of different objects can become very similar, resulting in unreliable appearances. 
Additionally, when an object remains in the same location and poses for a long period, its multiple appearances become identical, leading to redundancy.
We define the bounding box of the $m$-th object image $\mathbf{I}^m$ as $\text{bbox}(\mathbf{I}^m)=\left[x^m,y^m,w^m,h^m\right]$. 
We omit the camera indices $i$ and $j$ at this step for convenience.
In the video, the $m$-th object contains a set of appearances as
\begin{equation}
    \mathcal{A}^{m}=\{\mathbf{I}^{m,1}, \mathbf{I}^{m,2}, ... , \mathbf{I}^{m,N_m}\},
\end{equation}
where $N_m$ is the total number of appearances in $m$-th object.

\begin{figure}
	\centering
	\begin{subfigure}{0.9\linewidth}
		\includegraphics[width=\linewidth]{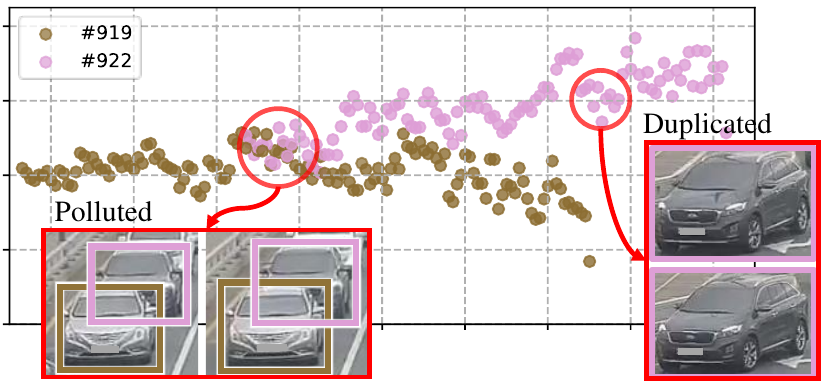}
		\caption{Without appearance management}
		\label{subfig:scatter001}
	\end{subfigure}
	\hfill
	\begin{subfigure}{0.9\linewidth}
		\includegraphics[width=\linewidth]{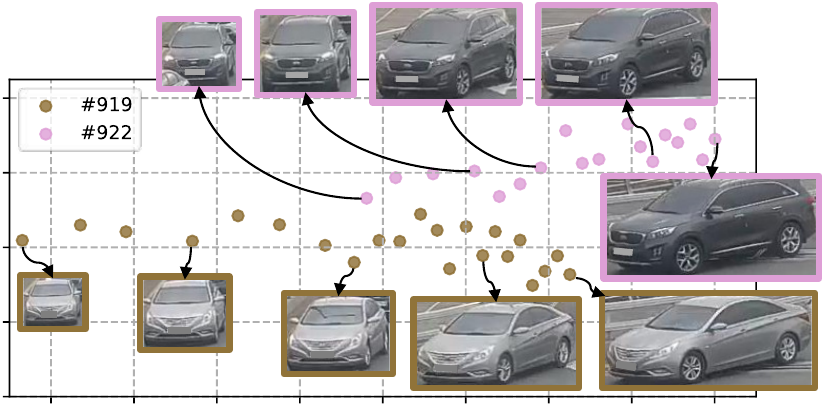}
		\caption{After appearance management}
		\label{subfig:scatter049}
	\end{subfigure}
	\caption{Feature distributions of objects. Example image patches reflect their actual sizes.}
	\label{fig:scatter}
\end{figure}

Inspired by \cite{cho2023rethinking}, we perform appearance management for each object.
To detect occluded unreliable objects, the occlusion ratio between two different objects is calculated based on Intersection over Union~(IoU) as $\text{IoU}\left(\mathbf{I}^m,\mathbf{I}^n\right)=\frac{\text{bbox}(\mathbf{I}^m)\cap\text{bbox}(\mathbf{I}^n)}{\text{bbox}(\mathbf{I}^m)\cup\text{bbox}(\mathbf{I}^n)}$.
In general, surveillance cameras are installed at a high position from the ground, looking down at an angle.
This means that an overlapping object with a larger $y$-value is occluding another object.
Therefore, we designed the occlusion ratio for the $m$-th object as
\begin{equation}
	Occ(\mathbf{I}^m,\mathbf{I}^n) =  
 \begin{cases}
     \text{IoU}\left(\mathbf{I}^m,\mathbf{I}^n\right),  & \text{if }  y^m < y^n \\
     0, & \text{otherwise}
 \end{cases}	
\end{equation}
where $m \neq n$.
When multiple bounding boxes occlude $\mathbf{I}^m$, the largest value will be used as the final occlusion ratio.

Next, we measure a self-overlapping ratio to prevent duplicate appearances from being included in the set $\mathcal{A}^m$.
For simplicity, let $\mathbf{\hat{I}}^m \in \mathcal{A}^m$ represent the last appearance in the set $\mathcal{A}^m$.
Then, the self-overlapping ratio is calculated using $\text{IoU}\left(\mathbf{I}^m,\mathbf{\hat{I}}^m\right)$.
Finally, we merge the two ratios to define the appearance unreliability score of $\mathbf{I}^m$ by
\begin{equation}
	\mathcal{U}\left(\mathbf{I}^m\right) = Occ(\mathbf{I}^m, \mathbf{I}^n) \cdot \text{IoU}\left(\mathbf{I}^m,\mathbf{\hat{I}}^m\right).
\end{equation}
It reflects both appearance occlusion and duplication.
Based on this, we can easily reject unreliable appearances by selecting only those with the lowest scores.

Figure~\ref{fig:scatter} illustrates the feature distributions of four different objects.
Due to occlusions, these objects are not separated in the feature space without the proposed appearance management.
Furthermore, each object has similar and redundant appearances when standing at the traffic signal. 
After performing appearance management, the unreliable appearances were appropriately removed, preserving the original distributions with a much smaller number of appearances.

\subsection{Top-$k$ Multi-shot Matching}
\label{sec:proposed_4}
After appearance management, objects contain more reliable appearances in their appearance sets.
To compare two different appearance sets, many traditional multi-shot matching ReID methods have tried maximum~(max) and average~(avg.) aggregation approaches.
However, the `max' approach, which selects only the maximum similarity value from multiple appearance comparisons, can produce highly biased final results.
On the other hand, the `avg.' approach, which averages all comparison results, can result in overly smoothed final outcomes.

To mitigate these risks, we perform Top-$k$ multi-shot matching to compare two different appearance sets.
Given two appearance sets $\mathcal{A}^m,\mathcal{A}^n$ with cardinalities $|\mathcal{A}^m| = N_m$, $|\mathcal{A}^n| = N_n$, respectively, there are $N_m \times N_n$ possible appearance comparisons.
Using the FusionNet described in Section~\ref{sec:proposed_2}, the appearance similarities between the two sets are represented by
\begin{equation}
	S_F\left(\mathcal{A}^m,\mathcal{A}^n\right) = \{S_F\left(\mathbf{I}^{m,1},\mathbf{I}^{n,1}\right),...,S_F\left(\mathbf{I}^{m,N_m},\mathbf{I}^{n,N_n}\right)\},
\end{equation}
Sort $S_F\left(\mathcal{A}^m,\mathcal{A}^n\right) $ in descending order and obtain
\begin{equation}
	S_{F_{sorted}}\left(\mathcal{A}^m,\mathcal{A}^n\right) = \{S_{F_1}, S_{F_2}, ... , S_{F_{N_m\times N_n}}\},
\end{equation}
Finally, the top-$k$ multi-shot matching is calculated by
\begin{equation}
    S_{F_{topk}}\left(\mathcal{A}^m,\mathcal{A}^n\right) = \frac{1}{k} \sum_{l=1}^k S_{F_l}.
\end{equation}
By adjusting the value $k$, we can control how many results are used in the multi-shot appearance comparison.
We empirically set $k$ to around $5$, which resulted in improved ReID performances, as shown in Table.~\ref{tab:result_2}.
The proposed top-$k$ approach is quite simple but effective.
It does not require complex structures for handling multi-shot comparisons like video-based ReID methods~\cite{li2019multi,zhang2021spatiotemporal} but can utilize single-shot baselines.
In addition, it implements other approaches by setting $k=1$ for the `max' approach and $k=N_m\times N_n$ for the `avg.' approach.

\subsection{Causal Identity Matching~(CIM)}
\label{sec:proposed_5}
In Section~\ref{sec:proposed_1}, the camera network topology was built as $\mathbf{G} = (\mathbf{V}, \mathbf{E})$, and FusionNet was trained based on this topology.
The topology includes transition time distributions $p_{ij}$ between all possible camera pairs, regardless of their connection strengths, allowing FusionNet to learn these strengths to determine the similarity $S_F$.
In real-world scenarios, we focus more on the target object (i.e., query) and its causality in the camera networks $\mathbf{G}$.
The causality indicates where the target object is currently and where it is likely to move next based on the topology.
In this setting, the query and gallery sets should not be fixed or pre-defined but should dynamically change according to the ReID results at each step.

To consider aforementioned conditions and perform causal identity matching~(CIM), we first define the camera adjacent matrix by $ \mathbf{C} \in \{c_{ij}|1\leq i\leq N_{cam}, 1\leq j\leq N_{cam}\}$, where $i,j$ are camera indexes, and $c_{ij}$ denotes a binary adjacent flag between cameras.
When cameras $c_{i}$ and $c_{j}$ are adjacent, the flag is set to $c_{ij}=1$; otherwise, it is set to $c_{ij}=0$. 
All diagonal elements of $\mathbf{C}$ are set to $c_{ii}=0$.
Given multiple object paths, we can simply find adjacent cameras and their overall connections. 
For example, assuming there are multiple object paths such as $\left(c_{1} \rightarrow c_{2} \rightarrow c_{3}\right)$, $\left(c_{2} \rightarrow c_{3} \right)$, $\left(c_{2} \rightarrow c_{4}\right)$, and $\left(c_{4} \rightarrow c_{2}\right)$; the camera adjacent matrix is defined by
\begin{equation}
	\small
	\mathbf{C} = 
	\begin{bmatrix}
		0 & 1 & 0 & 0 \\
		0 & 0 & 1 & 0 \\
		0 & 0 & 0 & 1 \\
		0 & 1 & 0 & 0
	\end{bmatrix}.
\end{equation}
Note that the matrix can be asymmetric when considering a bidirectional camera network topology.
More reliable adjacent connections can be estimated by voting on sufficient numbers of paths.
Additionally, unnecessary paths can be removed through post-processing algorithms like transitive reduction~\cite{aho1972transitive}\footnote{When there is a new path $\left(c_{1} \rightarrow c_{3} \right)$, it does not assign $c_{13}=1$, immediately.
Instead, the algorithm tries to infer if there is an intermediate vertex $c_2$ between $c_1$ and $c_3$ based on numerous training paths.}.
Excluding redundant paths helps to improve the overall efficiency of ReID.

\begin{figure*}
	\centering
	\begin{subfigure}{0.50\linewidth}
		\includegraphics[width=\linewidth]{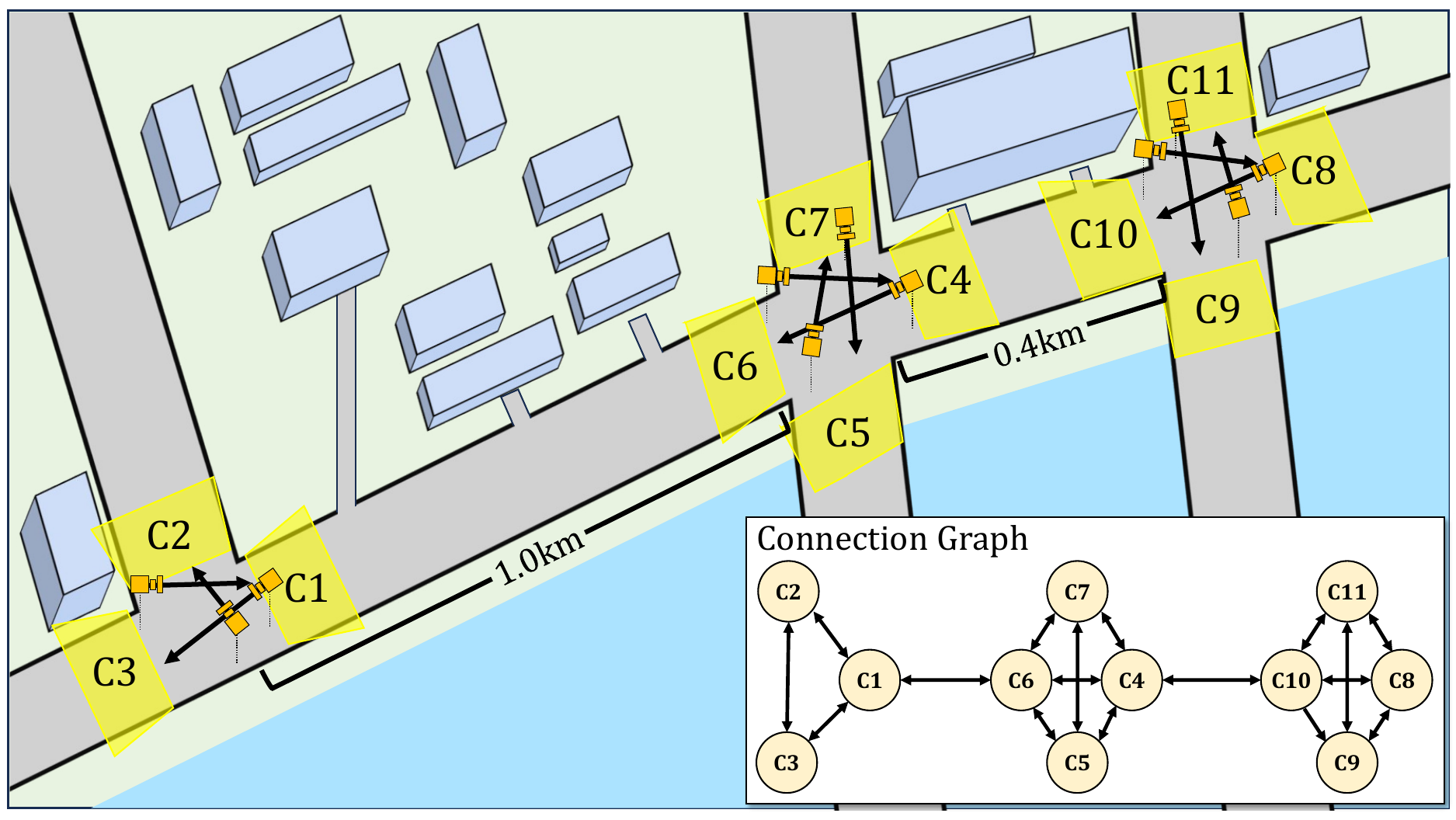}
		\caption{Layout and connection graph}
		\label{subfig:datset-map}
	\end{subfigure}
	\hfill
	\begin{subfigure}{0.44\linewidth}
		\includegraphics[width=\linewidth]{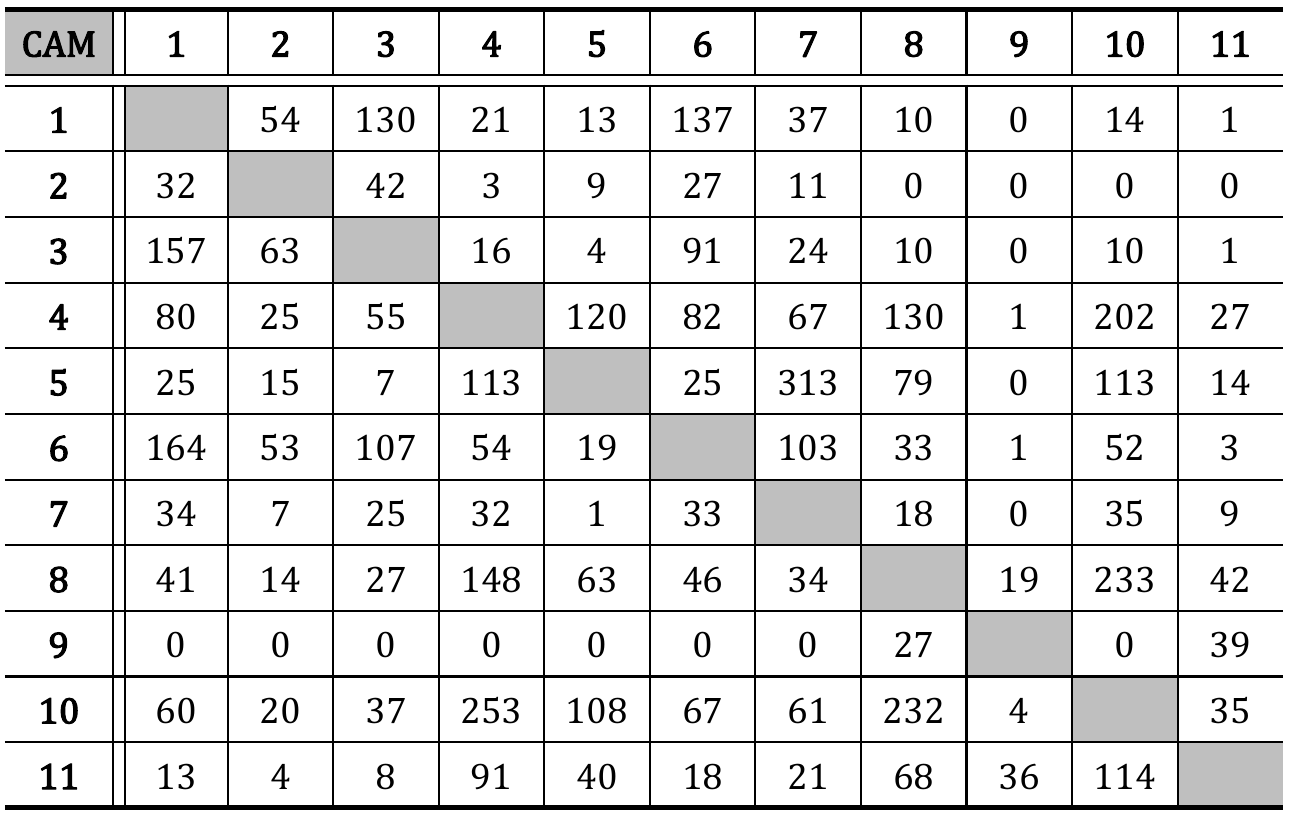}
		\caption{Number of vehicle pairs between cameras}
		\label{subfig:dataset-pair}
	\end{subfigure}
	\caption{Overall connections and vehicle flows in the \texttt{Vehicle-3I}}
	\label{fig:dataset-3i}
\end{figure*}

Next, we assign dynamic gallery sets for the target object.
Suppose a target object~(query) is detected by camera $c_1$.
Then, the gallery set of the query comes from the cameras adjacent to $c_1$ by finding $(c_{1j}=1)$ in $\mathbf{C}$.
Furthermore, the objects in the gallery are verified by referring to the corresponding transition time distributions $p_{1j}\left(t\right)$. 
Only objects within the camera $j$ that appear within a time $t$ satisfying $p_{1j}\left(t\right)>0$ are included in the gallery.
Through this process, we can causally assign the gallery set using $\mathbf{C}$ and reduce many unnecessary matching candidates based on $p_{ij}\left(t\right)$.

An object in the gallery with the highest similarity to the query is considered a successful match, and the query is merged with the re-identified object.
This increases the diversity of the query's appearance, making the subsequent ReID steps more advantageous.
In other words, although a single camera can only capture fixed poses of the object, progressive query merging easily overcomes this limitation.
The updated query is then used to progressively perform ReID on the following adjacent cameras, with the gallery adaptively determined by $\mathbf{C}$ and $p_{ij}$.

\section{Datasets}
\label{sec:datasets}

\begin{figure}
	\centering
	\includegraphics[width = 0.95
 \linewidth]{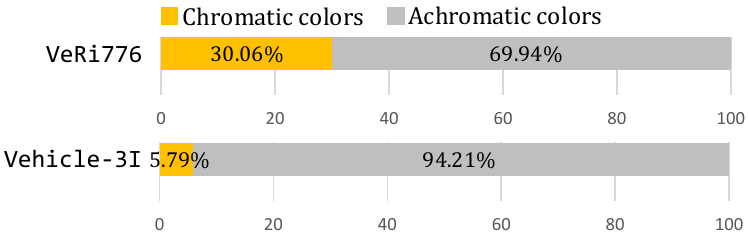}
	\caption{The ratio of achromatic to chromatic colors in the \texttt{VeRi776}~\cite{liu2016deep} and \texttt{Vehicle-3I} datasets is provided. White, gray, and black are categorized as achromatic colors, while all other colors are chromatic.}
	\label{fig:compare_dataset}
\end{figure}

For various experiments, we used \texttt{VeRi776}~\cite{liu2016deep} vehicle re-identification~(ReID) datasets, and \texttt{Market-1501}~\cite{zheng2015scalable} person ReID dataset.
In addition, to validate the proposed methods in a real-world scenario, we newly constructed the \texttt{Vehicle-3I} vehicle ReID dataset.
\begin{itemize}
	\item \texttt{VeRi776}~\cite{liu2016deep} dataset contains over 49,000 images of 776 different vehicle identities~(IDs) captured by 20 non-overlapping synchronized cameras. 
Each vehicle passes through 2 to 18 cameras with various viewpoints, illuminations, resolutions, and occlusions.
    \item \texttt{Vehicle-3I} is our new dataset for vehicle ReID, as illustrated in Fig.~\ref{fig:dataset-3i}. 
It includes data from $11$ synchronized cameras capturing vehicles at three intersections.
The dataset contains over 40,000 images with 2,038 IDs. 
These three intersection scenarios reflect the complexity and challenges of real-world traffic environments.
Unlike public datasets~\cite{zheng2015scalable, liu2016deep} that provide only cropped image patches, the \texttt{Vehicle-3I} dataset offers full-frame videos.
Furthermore, compared to the \texttt{VeRi776}~\cite{liu2016deep}, most vehicles in \texttt{Vehicle-3I} are achromatic ($94.21\%$), which poses a significant challenge for ReID (Fig.~\ref{fig:compare_dataset}).
Achromatic colors without hue, such as white, gray, and black, are difficult to discriminate under varying illumination conditions.

    \item \texttt{Market-1501}~\cite{zheng2015scalable} dataset contains over 32,000 images of 1,501 different person IDs captured by six non-overlapping synchronized cameras. Each image includes object IDs, timestamps~(frame No.), and camera IDs.
\end{itemize}

\section{Experimental Results}
\label{sec:exp}

\begin{table*}[t]
	\fontsize{8.2}{12}\selectfont
	\centering
	\renewcommand{\arraystretch}{0.85} 
	\setlength\tabcolsep{3.5pt}  
	\begin{tabular}{l|c|c|c|c|c|c|c|c|c|c|c|c}
		\noalign{\hrule height 1pt}
		\multirow{2}{*}{\textbf{Methods}} & \multicolumn{4}{c|}{\texttt{VeRi776}~\cite{liu2016deep}} & \multicolumn{4}{c|}{\texttt{Vehicle-3I}} & \multicolumn{4}{c}{\texttt{Market-1501}~\cite{zheng2015scalable}} \\ \cline{2-13}
		& \textbf{No. of IDs} & \textbf{Rank-1} & \textbf{Rank-5} & \textbf{mAP}
		& \textbf{No. of IDs} & \textbf{Rank-1} & \textbf{Rank-5} & \textbf{mAP}
		& \textbf{No. of IDs} & \textbf{Rank-1} & \textbf{Rank-5} & \textbf{mAP} \\ \hline
		
		\textit{FastReID}~\cite{he2020fastreid} &  576 for  & 96.96 & 98.45 & 81.91 
		& 1,630 for & 63.73 & 77.70 & 64.24
		& 751 for & 96.35 & - & 90.77  \\ \cline{1-1} \cline{3-5} \cline{7-9} \cline{11-13}
		\textit{FastReID}~\cite{he2020fastreid}+\textit{Wang's}~\cite{wang2019spatial}  
		& Appearance & 95.77 & 97.74 & 85.47 
		& Appearance & 81.37 & 88.60 & 66.34 
		& Appearance & 95.52 & 99.26 & 89.06 \\ \hline\hline
		
		FusionNet $(W=0)$  &  & 97.08 & 98.57 & 80.58 
		& & 54.41 & 77.08 & 55.13
		& & 95.52 & 98.22 & 90.08  \\ \cline{1-1} \cline{3-5} \cline{7-9} \cline{11-13}
		FusionNet $(W=2)$  & 526 for & 98.09 & 98.87 & 84.68 
		& 1,430 for & 86.40 & 92.52 & 80.37
		& 701 for & 95.84 & 99.29 & 92.09  \\ \cline{1-1} \cline{3-5} \cline{7-9} \cline{11-13}
		FusionNet $(W=4)$  & Appearance & 99.52 & 99.70 & 90.77 
		& Appearance & 87.87 & 93.38 & 82.40
		& Appearance & 98.52 & 99.44 & 92.32 \\ \cline{1-1} \cline{3-5} \cline{7-9} \cline{11-13}
		FusionNet $(W=6)$  & |||| & 99.64 & 99.82 & 91.45 
		& |||| & $\mathbf{89.22}$ & $\mathbf{93.50}$ & $\mathbf{83.62}$
		& |||| & 98.60 & 99.38 & 92.47  \\ \cline{1-1} \cline{3-5} \cline{7-9} \cline{11-13}
		FusionNet $(W=8)$  & 50 for & 99.64 & 99.82 & 91.55 
		& 200 for & 87.75 & 92.52 & 82.66 
		& 50 for & 97.27 & 99.52 & 92.83 \\ \cline{1-1} \cline{3-5} \cline{7-9} \cline{11-13}
		FusionNet $(W=10)$ & FusionNet & \textbf{99.70} & \textbf{99.82} & \textbf{91.71} 
		& FusionNet & 88.85 & 92.65 & 83.36 
		& FusionNet & $\mathbf{99.11}$ & $\mathbf{99.58}$ & $\mathbf{93.80}$ \\ \cline{1-1} \cline{3-5} \cline{7-9} \cline{11-13}
		FusionNet $(W=12)$ &  & 99.64 & 99.82 & 91.57 
		& & 88.48 & 93.01 & 82.91 
		& & 98.84 & 99.52 & 93.52 \\ \noalign{\hrule height 1pt}
	\end{tabular}
	\caption {ReID performances according to different $W$ values in FusionNet. 
		The \textit{FastReID}~\cite{he2020fastreid} method does not utilize spatial-temporal information $S_T$. 
		\textit{FastReID+Wang's} method estimates the final similarity by $S_F=S_A(\textbf{I}^m_i,\textbf{I}^n_j) \cdot p_{ij}\left(\tau^{mn}_{ij}\right)$, as in~\cite{wang2019spatial} and employed \textit{FastReID}~\cite{he2020fastreid} for its appearance model. The best performances are marked in \textbf{bold}.}
	\label{tab:1}
\end{table*}

\subsection{Settings}
\label{sec:setting}
To estimate the camera network topology $\bold G = (\bold V, \bold E)$, we used training datasets containing ~(576 IDs from \texttt{VeRi776}, 1,630 IDs from \texttt{Vehicle-3I}, and 751 IDs from \texttt{Market-1501}).
Of these datasets, 90\% of the training data was used for appearance-based ReID model training, while the remaining 10\% was reserved for FusionNet training. It is important to note that the object identities~(IDs) were completely separated for each training task.

In \texttt{VeRi776}, with 20 cameras~($N_{cam}=20$), the estimated camera network topology contains 400 object transition time distributions $(p_{ij})$. 
Of these, 380 distributions are between different camera pairs (i.e., $p_{ij}$, where $c_i \neq c_j$), with each distribution consisting of 300 bins, each covering a range of 100 frames.
All distributions were estimated using the proposed adaptive Parzen window method.  
The same processes were applied to estimate the camera network topology for \texttt{Vehicle-3I} and \texttt{Market-1501}.

For the appearance-based ReID model in our framework, we trained FastReID~\cite{he2020fastreid} and SOLIDER~\cite{chen2023beyond} to extract appearance similarity.
The hyperparameters for FastReID across both datasets were as follows: epoch —- 60, batch size -— 64.   
For the vehicle ReID task, a simple ResNet-50~\cite{he2016deep} was utilized as the backbone network structure of FastReID.
Meanwhile, for the Person ReID task, ResNet-101 with MGN~\cite{wang2018learning} was utilized as the backbone network structure of FastReID.

The proposed FusionNet has a single hidden layer, with the number of nodes in the hidden layer set to approximately 65\% of the input vector size, calculated as $(2(2W+2)/3+1)$ and rounded accordingly. 
The training parameters for FusionNet are as follows: epoch —- 40, batch size —- 128, learning rate —- 0.001, optimizer —- Adam. 
The datasets \texttt{VeRi776}~\cite{liu2016deep}, \texttt{Vehicle-3I} and \texttt{Market-1501}~\cite{zheng2015scalable} contain 200, 408, and 750 different IDs, respectively, for method validation and performance evaluation.
We evaluated rank-$1$ and rank-$5$ accuracy, and as mAP, based on the traditional 1-to-all single-shot comparison, as illustrated in Fig.~\ref{fig:single_multi_matching}.

\subsection{Effects of FusionNet}
\label{sec:fusionnet}

In this experiment, we tested the effect of the $W$ value on FusionNet using the \texttt{VeRi776}~\cite{liu2016deep}, \texttt{Market-1501}~\cite{zheng2015scalable}, and \texttt{Vehicle-3I} datasets.
The results are presented in Table~\ref{tab:1}. 
Additionally, we compared other methods, such as \textit{FastReID}~\cite{he2020fastreid}, which uses only appearance similarities~$(S_A)$ between images. \textit{FastReID} + \textit{Wang's} method estimates the appearance similarity~$(S_A)$ based on FastReID~\cite{he2020fastreid} and combines it with spatial-temporal similarity using $S_F=S_A(\textbf{I}^m_i,\textbf{I}^n_j) \cdot p_{ij}\left(\tau^{mn}_{ij}\right)$, as described in~\cite{wang2019spatial}.

In the \texttt{VeRi776}~\cite{liu2016deep}, the \textit{FastReID}~\cite{he2020fastreid} model, which uses only appearance information, achieved a rank-1 accuracy of 96.96\% and an mAP of 81.91
However, combining \textit{FastReID} with \textit{Wang's} approach, which merges additional spatial-temporal similarity, improved the mAP of the appearance-based ReID~\cite{he2020fastreid} by 3.56\%. 
Despite this improvement in mAP, the simple combination of these two different similarities using $S_A(\textbf{I}^m_i,\textbf{I}^n_j) \cdot p_{ij}\left(\tau^{mn}_{ij}\right)$ is not yet optimized. 
As a result, the rank-1 and rank-5 performances decreased after implementing \textit{Wang's} approach in \textit{FastReID}~\cite{he2020fastreid}.
Using the proposed FusionNet, we achieved a significant improvement in ReID performance. 
For fair comparisons, we used \textit{FastReID}~\cite{he2020fastreid} as the appearance model in our framework and trained FusionNet with 526 IDs for the appearance model and 50 IDs for FusionNet training.
With $W$ set to 10, FusionNet achieved the best performance in rank-1 accuracy at 99.70\%, rank-5 accuracy at 99.82\%, and mAP at 91.71\%. Except at $W=0$, FusionNet outperformed other methods in all evaluation metrics (rank-1, -5, and mAP).

In the \texttt{Vehicle-3I} dataset, \textit{FastReID}~\cite{he2020fastreid} using only appearance information, achieved a rank-1 accuracy of 63.73\% and an mAP of 64.24\%.
The rank-1, rank-5, and mAP performances increased after adopting \textit{wang's} approach in \textit{FastReID}~\cite{he2020fastreid}.
This indicates that using spatial-temporal information in the \texttt{Vehicle-3I} dataset is also effective. 
Except for $W=0$, our proposed method significantly outperformed \textit{FastReID}~\cite{he2020fastreid} and \textit{FastReID}~\cite{he2020fastreid}+\textit{Wang's}~\cite{wang2019spatial}. 
At $W=6$, we achieved the best performance with 89.22\% rank-1 accuracy, 93.50\% rank-5 accuracy, and 83.62\% mAP.

In the \texttt{Market-1501}~\cite{zheng2015scalable}, person ReID dataset, \textit{FastReID}~\cite{he2020fastreid} achieved a rank-1 accuracy of 96.35\% and an mAP of 90.77\% using only appearance information. 
However, the rank-1 accuracy and mAP decreased after adapting \textit{Wang's} approach into \textit{FastReID}~\cite{he2020fastreid}. 
Despite this, the proposed framework significantly enhanced ReID performance. 
FusionNet achieved the best performance with a rank-1 accuracy of 99.11\%, rank-5 accuracy of 99.58\%, and mAP of 93.8\% when $W$ was set to 10. 
These results demonstrate that the proposed FusionNet can optimally combine different types of information, such as appearance similarity and spatial-temporal information. 
Additionally, even with the reduced training images for the appearance model, it achieved superior performance compared to other methods.

 \begin{figure}
 	\centering
 	\includegraphics[width=0.8\linewidth]{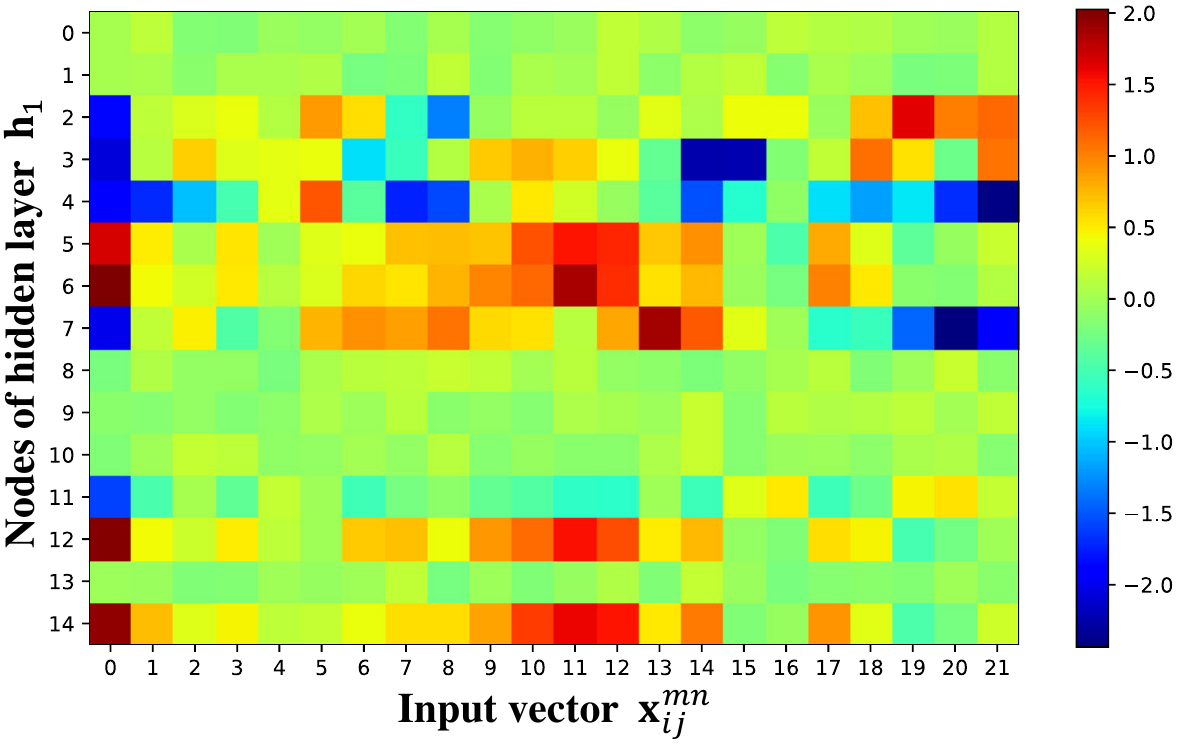}
 	\caption{A visualization of the trained weight vector~($\mathbf{w}_1$) between the input layer and the hidden layer~($\mathbf{h}_1$)}
 	\label{fig:4}
 \end{figure}

Fig.~\ref{fig:4} illustrates the trained weight vector~($\mathbf{w}_1$) between the input layer and the hidden layer~($\mathbf{h}_1$). 
The index of the nodes in $\mathbf{h}_1$ is denoted by the row numbers~(0–14), and the index of the input vector $\mathbf{x}^{mn}_{ij}$ is denoted by the column numbers~(0–21). 
The first column~(index 0) represents the weights for appearance similarity~($S_A$), which are notably more significant than other weights. This indicates that FusionNet has effectively learned the importance of appearance similarity~($S_A$). 
The weights in columns 1 to 21 correspond to the spatial-temporal distribution~($S_T$).
Interestingly, the weights in columns 10 to 12 exhibit large magnitudes. 
This result indicates that spatial-temporal information related to the time difference ($\tau^{mn}_{ij}$) between two images plays a key role in ReID.
Additionally, FusionNet has a lightweight yet effective structure.

We also tested the effect of hidden layer depth in FusionNet. 
We observed that a single hidden layer achieved the best mAP performance~(91.71\%). 
The performance decreased significantly to a 59.47\% mAP score with a simple perceptron without a hidden layer. 
This indicates that a simple perceptron structure is insufficient for effectively combining appearance similarity and spatial-temporal information. Interestingly, increasing the depth of hidden layers did not lead to an improvement in mAP score; in fact, the mAP scores tended to decrease: for example, 2-layer: 91.22\%, 3-layer: 91.28\%, and 4-layer: 91.22\%. 
It suggests that many hidden layers are unnecessary, and a single hidden layer is sufficient for learning the relationships between appearance similarity and spatial-temporal similarity. 

\begin{figure*}
  \centering
  \begin{subfigure}{0.325\linewidth}
    \includegraphics[width=\linewidth]{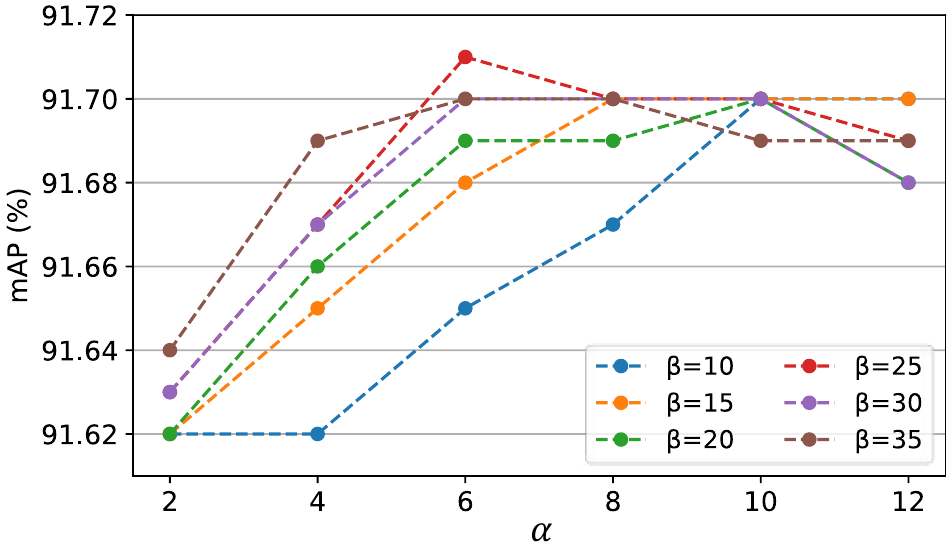}
    \caption{\texttt{VeRi776}~\cite{liu2016deep}}
    \label{subfig:sub_veri}
  \end{subfigure}
  \hfill
  \begin{subfigure}{0.325\linewidth}
    \includegraphics[width=\linewidth]{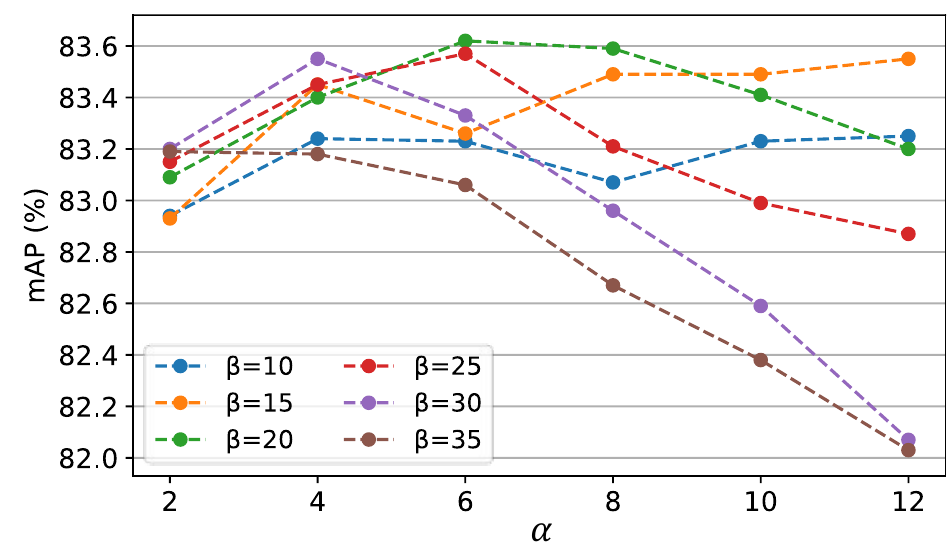}
    \caption{\texttt{Vehicle-3I}}
    \label{subfig:sub_etri}
  \end{subfigure}
  \hfill
  \begin{subfigure}{0.325\linewidth}
    \includegraphics[width=\linewidth]{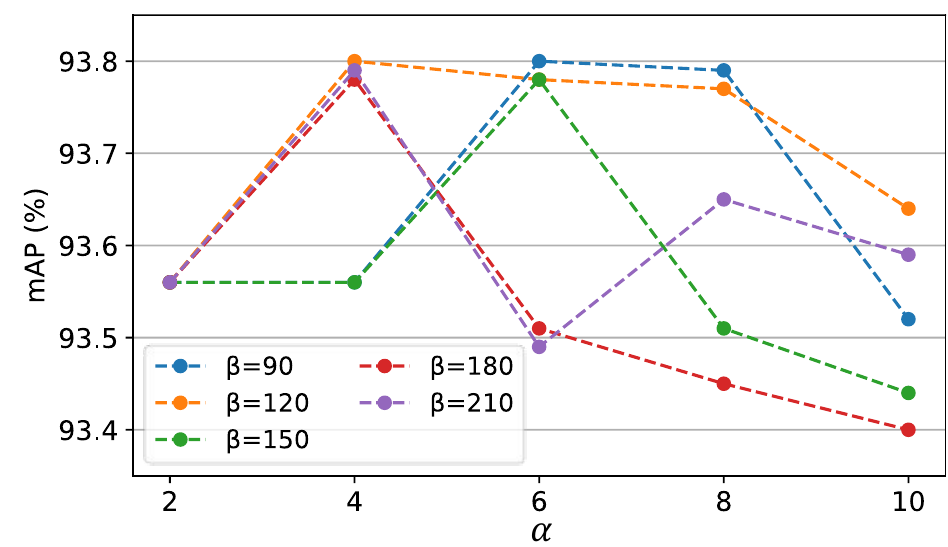}
    \caption{\texttt{Market-1501}~\cite{zheng2015scalable}}
    \label{subfig:sub_market}
  \end{subfigure}
  \caption{Performance according to the $\alpha$ and $\beta$ factors of the adaptive Parzen window.}
  \label{fig:alpha_beta}
\end{figure*}

\subsection{Effects of adaptive Parzen window}
\label{sec:parzen}

We evaluated various factors for the proposed adaptive Parzen window method, including the scale factor $\alpha$ and smoothness factor $\beta$ in Eq.~\ref{eq:4}. 
We set $W=10$ for FusionNet on the \texttt{VeRi776}~\cite{liu2016deep} and \texttt{Market-1501}~\cite{zheng2015scalable} datasets, and $W=6$ for FusionNet on the \texttt{Vehicle-3I} dataset.
Fig.~\ref{fig:alpha_beta} illustrates the mAP according to the $\alpha$ and $\beta$ factors. 
When the scale factor $\alpha=6$ and the smoothness factor $\beta=25$, the proposed framework achieved the best mAP performance on \texttt{VeRi776}~\cite{liu2016deep}. 
Similarly, the proposed framework achieved the best mAP performance with $\alpha=6, \beta=20$ on \texttt{Vehicle-3I} and $\alpha=4, \beta=120$ on \texttt{Market-1501}~\cite{zheng2015scalable}. 
Notably, ReID performances do not significantly change with variations in the factor values, as the difference between the maximum and minimum mAP is 0.09\%, 1.77\%, and 0.47\% for =\texttt{VeRi776}~\cite{liu2016deep}, \texttt{Vehicle-3I} and \texttt{Market-1501}~\cite{zheng2015scalable}.

We compared the ReID performances using fixed and the proposed adaptive $\sigma$ values. 
The fixed $\sigma=1,5,10,100$, and the adaptive $\sigma$ as in Eq.~\ref{eq:2} were tested. 
As summarized in Table~\ref{tab:2}, the fixed $\sigma=100$ performed the worst because a large $\sigma$ leads to overly smoothed distributions that approximate a uniform distribution, thereby reducing the influence of spatial-temporal information.
Conversely, $\sigma=1$, $\sigma=5$, and $\sigma=10$ performed relatively well.
However, compared to the fixed $\sigma$ values, the proposed adaptive Parzen window with the adaptive $\sigma$ achieved the best performance across all evaluation metrics~(Rank-1, mAP) for all tested datasets.
The proposed adaptive Parzen window achieved rank-1 accuracy of 99.70\%, 89.22\%, and 99.11\%, and mAP scores of 91.71\%, 83.62\%, and 93.80\% on three different datasets, respectively. 
As explained in Section~\ref{sec:proposed_1} and shown in Fig.~\ref{fig:three_figures}, a fixed $\sigma$ value struggles to handle various types of initial histograms $h_{ij}$. 
This result implies that setting different $\sigma_{ij}$ values by considering the varying connection strengths of camera pairs is effective and improves ReID performance.

\begin{table}[t]
    \small
    \centering
    \renewcommand{\arraystretch}{0.55} 
    \begin{tabular}{l|r|r|r}
    \noalign{\hrule height 1pt}
    Dataset & \multicolumn{3}{c}{\texttt{VeRi776}~\cite{liu2016deep}} \\ \hline
    Methods & Rank-1 & Rank-5 &  mAP \\ \hline
    Fixed $\sigma=1$         & $\mathbf{99.70}$ & $\mathbf{99.82}$ & 91.57 \\
    Fixed $\sigma=5$         & 99.52 & 99.70 & 91.43 \\
    Fixed $\sigma=10$        & 99.28 & 99.76 & 91.28 \\
    Fixed $\sigma=100$       & 98.81 & 99.46 & 86.62 \\ \hline
    Adaptive $\sigma$  (ours) & $\mathbf{99.70}$ & $\mathbf{99.82}$ & $\mathbf{91.71}$ \\ \hline \hline

    Dataset & \multicolumn{3}{c}{\texttt{Vehicle-3I}} \\ \hline
    Methods & Rank-1 & Rank-5 &  mAP \\ \hline
    Fixed $\sigma=1$         & 88.36 & $\mathbf{93.63}$ & 83.03 \\
    Fixed $\sigma=5$         & 88.60 & 93.38 & 82.96 \\
    Fixed $\sigma=10$        & 83.58 & 92.65 & 79.91 \\
    Fixed $\sigma=100$       & 69.73 & 80.88 & 68.25 \\ \hline
    Adaptive $\sigma$  (ours) & $\mathbf{89.22}$ & 93.50 & $\mathbf{83.62}$ \\ \hline \hline

    Dataset & \multicolumn{3}{c}{\texttt{Market-1501}~\cite{zheng2015scalable}} \\ \hline
    Methods & Rank-1 & Rank-5 &  mAP \\ \hline
    Fixed $\sigma=1$         & 99.02 & $\mathbf{99.58}$ & 93.56 \\
    Fixed $\sigma=5$         & 98.81 & 99.55 & 93.34 \\
    Fixed $\sigma=10$        & 98.25 & 99.41 & 92.12 \\
    Fixed $\sigma=100$       & 97.95 & 99.08 & 91.91 \\ \hline
    Adaptive $\sigma$  (ours) & $\mathbf{99.11}$ & $\mathbf{99.58}$ & $\mathbf{93.80}$ \\ 
    \noalign{\hrule height 1pt}
    \end{tabular}
\caption{ReID performance according to the values of $\sigma$ for Parzen window. The best performances are marked in \textbf{bold}.}
\label{tab:2}
\end{table}

\subsection{Comparison with state-of-the-art methods}
\label{sec:exp_4}

In this section, we compare the proposed method with state-of-the-art re-identification methods using the \texttt{VeRi776}~\cite{liu2016deep}, \texttt{Vehicle-3I} vehicle ReID dataset and \texttt{Market-1501}~\cite{zheng2015scalable} person ReID dataset. 
The methods are primarily categorized into two approaches: 
1) appearance-based only and 2) using additional spatial-temporal information~(marked by $\dagger$). 
The methods marked by a `*' performed re-ranking post-processing for the final ReID results. 

Table~\ref{tab:3} summarizes the vehicle ReID results, sorted by rank-1 performance.
Our framework achieved outstanding performance on the \texttt{VeRi776}~\cite{liu2016deep} dataset, with a rank-1 accuracy of 99.7\% and an mAP score of 91.71\%. 
Despite improvements to their baseline methods, the spatial-temporal approaches~\cite{liu2016deep, shen2017learning, liu2017provid} did not match the performance of other appearance-based methods due to their use of older appearance models such as SIFT, bag-of-words, and Siamese-CNN for ReID. 
Additionally, these methods did not directly estimate the camera network topology or optimize the use of spatial-temporal information.

Many appearance-based methods have significantly improved ReID performance by developing deep learning models. 
For instance, our baseline appearance model, FastReID~\cite{he2020fastreid}, achieved a 96.96\% rank-1 accuracy and 81.91\% mAP. 
RPTM~\cite{ghosh2023relation} notably used the GMS~\cite{bian2017gms} feature matcher and adopted the ResNet-101~\cite{he2016deep} structure, demonstrating superior performance with a 97.3\% rank-1 accuracy and an 88.0\% mAP.
In contrast, despite using a lightweight ResNet-50 structure for the appearance model, our methods outperformed the sophisticated RPTM~\cite{ghosh2023relation} approach, achieving a 2.4\% higher rank-1 accuracy, a 1.42\% higher rank-5 accuracy, and a 3.71\% higher mAP score. 
Furthermore, our methods did not perform any re-ranking processes for post-processing.

For \texttt{Vehicle-3I}, several state-of-the-art methods were evaluated. 
FastReID~\cite{he2020fastreid}, achieved a 63.73\% rank-1 accuracy and 64.24\% mAP.
In contrast, CLIP-ReID~\cite{li2023clip}, which achieved the second-best performance on \texttt{VeRi776}, showed the lowest performance on \texttt{Vehicle-3I} because it was trained without viewpoint annotations in this experiment.
Adopting \textit{Wang's}~\cite{wang2019spatial} approach to FastReID~\cite{he2020fastreid} improved performance to 81.37\% rank-1 accuracy and 88.60\% rank-5 accuracy, demonstrating the effectiveness of spatial-temporal cues.
Our method achieved outstanding results, outperforming all other methods with 89.22\% rank-1 accuracy and 83.62\% mAP. 
This result demonstrates that combining spatial-temporal information with appearance information is more effective than using appearance information alone in challenging conditions.

\begin{table}[]
	\centering
	\fontsize{8.2}{12}\selectfont
    \renewcommand{\arraystretch}{0.55} 
	\begin{tabular}{l|r|r|r}
		\noalign{\hrule height 1pt}
            Dataset & \multicolumn{3}{c}{\texttt{VeRi776}~\cite{liu2016deep}} \\ \hline
		Models  & Rank-1 & Rank-5 & mAP \\ \hline
		$\dagger$FACT+Plate-SNN+STR~\cite{liu2016deep} & 61.44 & 78.78 & 27.77 \\
		$\dagger$Siamese-CNN+Path-LSTM~\cite{shen2017learning} & 83.49 & 90.04 & 58.27 \\
		$\dagger$PROVID~\cite{liu2017provid}           & 81.56 & 95.11 & 53.42 \\ 
		$\dagger$KPGST~\cite{huang2022vehicle} & 92.35 &  93.92 & 68.73 \\
		$\dagger$FastReID~\cite{he2020fastreid} + Wang's~\cite{wang2019spatial} & 95.77 & 97.74 & 85.47 \\ \hline
		GAN+LSRO*~\cite{wu2018joint}            & 88.62 & 94.52 & 64.78 \\
		AAVER*~\cite{khorramshahi2019dual}      & 90.17 & 94.34 & 66.35 \\
		PAMTRI~\cite{tang2019pamtri}           & 92.86 & 96.97 & 71.88 \\
		SPAN~\cite{chen2020orientation}        & 94.00 & 97.60 & 68.90 \\
		CAL~\cite{rao2021counterfactual}       & 95.40 & 97.90 & 74.30 \\
		PVEN~\cite{meng2020parsing}            & 95.60 & 98.40 & 79.50 \\
		TBE~\cite{sun2021tbe}                  & 96.00 & \underline{98.50} & 79.50 \\
		TransReID~\cite{he2021transreid}       & 96.90 & - & 80.60 \\ 
		VehicleNet*~\cite{zheng2020vehiclenet}  & 96.78 & - & 83.41 \\
		SAVER*~\cite{khorramshahi2020devil}     & 96.90 & 97.70 & 82.00 \\
		DMT*~\cite{he2020multi}                 & 96.90 & - & 82.00 \\
		FastReID~\cite{he2020fastreid} & 96.96 & 98.45 & 81.91 \\
            CLIP-ReID~\cite{li2023clip}    & \underline{97.30} & - & 84.50 \\
		Strong Baseline*~\cite{huynh2021strong} & 97.00 & - & 87.10 \\
		RPTM*~\cite{ghosh2023relation}          & \underline{97.30} & 98.40 & \underline{88.00} \\ \hline
		$\dagger$\textbf{Ours}~(Appearance--FastReID~\cite{he2020fastreid})    & $\mathbf{99.70}$ & $\mathbf{99.82}$ & $\mathbf{91.71}$ \\ \hline \hline

        Dataset & \multicolumn{3}{c}{\texttt{Vehicle-3I}} \\ \hline
        Models  & Rank-1 & Rank-5 & mAP \\ \hline
        $\dagger$FastReID~\cite{he2020fastreid} + Wang's~\cite{wang2019spatial} & \underline{81.37} & \underline{88.60} & \underline{66.34} \\ \hline
        CLIP-ReID~\cite{li2023clip} & 61.90 & 68.80 & 59.00 \\
        FastReID~\cite{he2020fastreid} & 63.73 & 77.70 & 64.24 \\ \hline
        $\dagger$\textbf{Ours}~(Appearance--FastReID~\cite{he2020fastreid}) & $\mathbf{89.22}$ & $\mathbf{93.50}$ & $\mathbf{83.62}$ \\ 
        \noalign{\hrule height 1pt}
	\end{tabular}
\caption {Performance comparisons on Vehicle ReID datasets. $\dagger$ and $*$ indicate the spatial-temporal approach and re-ranking, respectively. The best and second best performances are marked in \textbf{bold} and \underline{underline}.}
\label{tab:3}
\end{table}

\begin{table}[]
	\centering
	\fontsize{8.2}{10}\selectfont
    \setlength\tabcolsep{6.0pt}  
    \renewcommand{\arraystretch}{0.55} 
	\begin{tabular}{l|r|r|r}
		\noalign{\hrule height 1pt}
		\textbf{Models}  & \textbf{Rank-1} & \textbf{Rank-5} & \textbf{mAP} \\ \hline
		GCP~\cite{park2020relation}        & 95.2 & - & 88.9 \\
        TransReID~\cite{he2021transreid}   & 95.2 & - & 89.5 \\
        ISP~\cite{zhu2020identity}         & 95.3 & 98.6 & 88.6 \\ 
        GASM~\cite{he2020guided}           & 95.3 & - & 84.7 \\
        ABDNET~\cite{chen2019abd}          & 95.6 & - & 88.28 \\
        SCSN~\cite{chen2020salience}       & 95.7 & - & 88.5 \\
        CLIP-ReID~\cite{li2023clip}        & 95.7 & - & 89.8 \\
		SAN~\cite{jin2020semantics}        & 96.1 & - & 88.0 \\
        FastReID~\cite{he2020fastreid}     & 96.35 & - & 90.77 \\
        PASS~\cite{zhu2022pass}            & 96.9 & - & 93.3 \\
        SOLIDER~\cite{chen2023beyond}      & 96.9 & - & \underline{93.9} \\
		Unsupervised Pre-training~\cite{fu2021unsupervised}    & 97.0 & - & 92.0 \\ 
		UP-ReID~\cite{yang2022unleashing}  & 97.1 & - & 91.1 \\ 
        $\dagger$st-ReID~\cite{wang2019spatial} & 98.1 & 99.3 & 87.6 \\  \hline
		$\dagger$\textbf{Ours} (Appearance--FastReID~\cite{he2020fastreid}) & $\mathbf{99.11}$ & \underline{99.58} & 93.8 \\ 
        $\dagger$\textbf{Ours} (Appearance--SOLIDER~\cite{chen2023beyond}) & \underline{99.0} & $\mathbf{99.6}$ & $\mathbf{95.5}$ \\ \noalign{\hrule height 1pt}
	\end{tabular}
\caption {Performance comparisons on the~\texttt{Market- 1501}~\cite{zheng2015scalable} Person ReID dataset. $\dagger$ indicate the spatial-temporal approach, respectively. The best and second best performances are marked in \textbf{bold} and \underline{underline}.}
\label{tab:4}
\end{table}

\begin{figure*}
	\centering
	\includegraphics[width=0.9\linewidth]{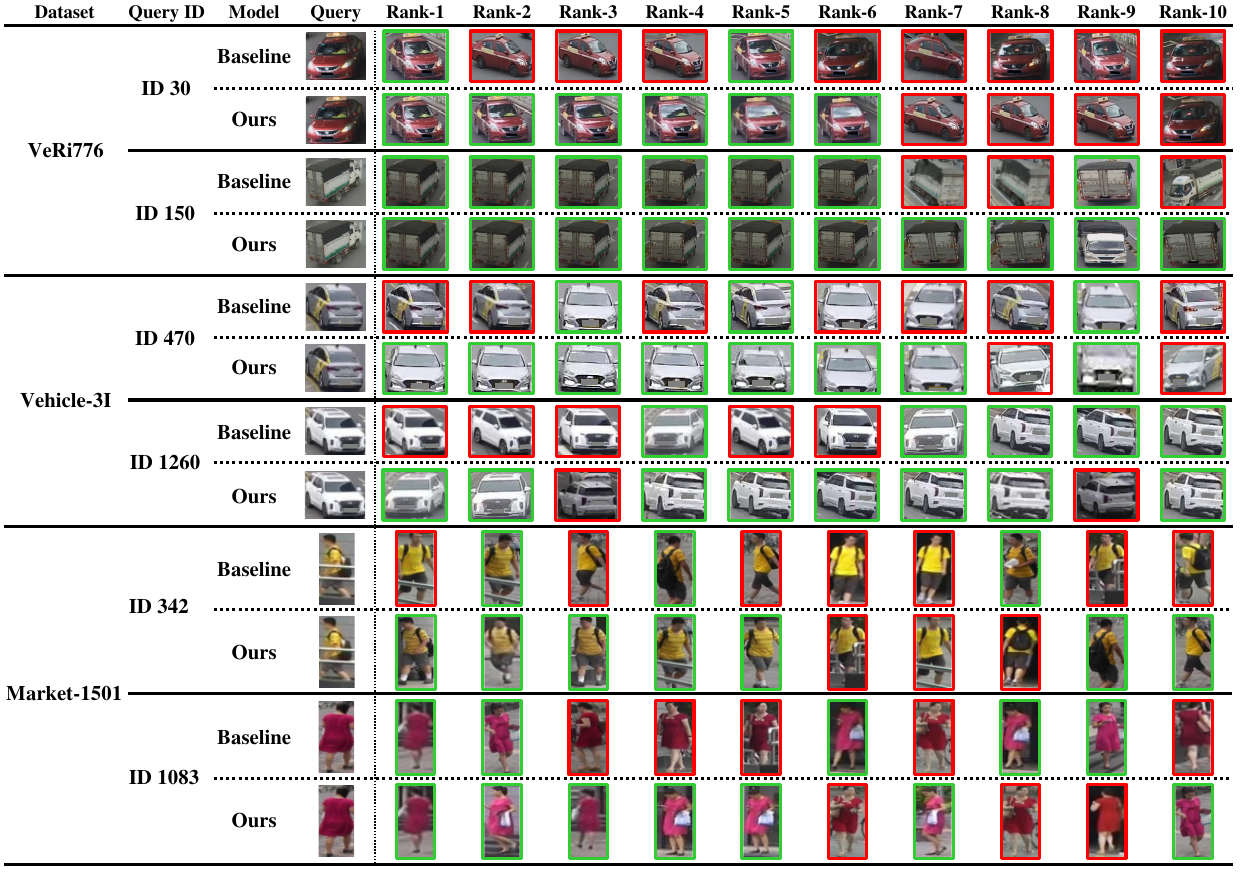}
	\caption{Qualitative ReID results of the baseline and proposed methods~(Ours). The baseline model~\cite{he2020fastreid} is an appearance-based method. Green and red boxes denote true and false matching. Compared to the baseline method, the proposed method can match true positive pairs despite similar appearances and overcome the appearance ambiguities due to the spatial-temporal information (best viewed in color).}
	\label{fig:6}
\end{figure*}

The Person ReID results from the \texttt{Market-1501}~\cite{zheng2015scalable} dataset are summarized in Table~\ref{tab:4}. 
Despite the st-ReID~\cite{wang2019spatial} achieving a rank-1 accuracy of 98.1\%, its mAP of 87.6\% was lower than that of other state-of-the-art methods. 
In contrast, SOLIDER~\cite{chen2023beyond}, which uses the Swin transformer~\cite{liu2021swin}, performed well with an mAP of 93.9\%, though its rank-1 accuracy was relatively lower at 96.9\%..
For this dataset, we selected two different appearance-based models, FastReID~\cite{he2020fastreid} and SOLIDER~\cite{chen2023beyond}, as baselines for our framework. 
Our frameworks achieved the best rank-1 accuracy (99.11\%) with FastReID~\cite{he2020fastreid}, and the best rank-5 accuracy (99.6\%) and mAP (95.5\%) with SOLIDER~\cite{chen2023beyond}. 
These results imply that the proposed spatial-temporal framework improves both vehicle and person ReID tasks and has the potential to achieve higher performance when it employs a better appearance-based model as its baseline.

Fig.~\ref{fig:6} illustrates the qualitative ReID results.  
We compared the proposed method~(Ours) with the baseline method (FastReID~\cite{he2020fastreid}), which used only appearance information.
The baseline method shows numerous false matches where the appearance closely resembles that of the query images. 
For instance, the baseline incorrectly matched ID 470 and 1260 images in the \texttt{Vehicle-3I}, which have the same color and model type. 
In contrast, the proposed method correctly matched most of the same vehicles.
Notably, it rarely matched the correct images of the 662-nd vehicle query image in \texttt{VeRi776}~\cite{liu2016deep} despite many similar black cars.  
The proposed method, leveraging spatial-temporal information, perfectly matched the correct images at rank-1 to rank-10 under the challenging query and gallery pairs. 
Conversely, the baseline model struggled to match correct pairs in the gallery during the person ReID task. 
For example, ID 342, wearing a stripe yellow shirt, gray shorts, and a black backpack, was easily confused with others in similar outfits.
Similarly, ID 1083, wearing a red dress and carrying a shoulder bag, was often mistaken for another person wearing a similar red dress with a handbag.

\begin{figure}
	\centering
	\includegraphics[width = 0.85 \linewidth]{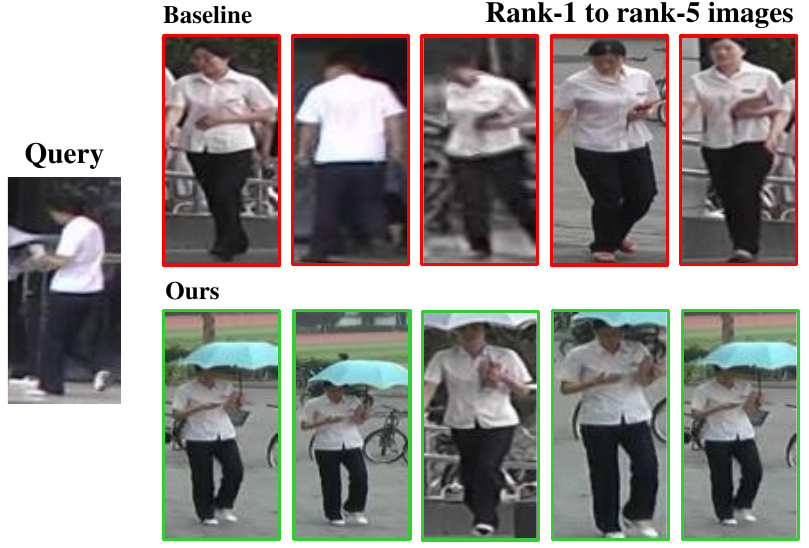}
	\caption{A challenging scenario in ReID. Green and red boxes denote true and false matches. In this case, a query image did not show the subject carrying an umbrella, whereas the subject carried an umbrella in other images captured by different cameras~(gallery). The baseline method (\textit{FastReID}~\cite{he2020fastreid}), which relies solely on appearance information, failed the correct match in the gallery. In contrast, the proposed methods successfully identified the true match.}
	\label{sup_fig:qualitative_person}
\end{figure}

Interestingly, as shown in Fig.~\ref{sup_fig:qualitative_person}, the query image depicts a person wearing a white shirt and black pants, but not carrying an umbrella at that time.
Based on the appearance model, it naturally found the images with white shirts and black pants in the gallery.
However, the query carried an umbrella in other cameras a few minutes later.
In this challenging ReID scenario, the appearance-based model, relying solely on visual cues, failed to find the correct matches.
In contrast, our method, which incorporates spatial-temporal information, correctly matched the images despite substantial changes in the query’s appearance.
These results demonstrate that the proposed ReID method, based on the spatial-temporal fusion network, effectively manages appearance ambiguity and overcomes the limitations of previous ReID methods.

\section{ReID in Real-world Surveillance Scenarios}
\label{sec:exp_5}
\subsection{Settings and evaluation metrics}
Unlike the appearance-centric~(1-to-all) evaluation settings in Section.~\ref{sec:exp}, in this experiment, we perform ID-to-ID evaluation to consider the real-world scenarios. 
This means that ReID performances should be evaluated only once for each identity between cameras. 
For example, each identity has a path across cameras, e.g., $c_1\rightarrow c_2\rightarrow c_3 \rightarrow c_4$; thus, ReID methods should correctly track the path based on the ReID between adjacent cameras.
In real-world ReID scenarios, only the rank-1 matching result is considered valid. Therefore, we regarded the rank-1 result with a similarity value over $\theta_c$ as a positive match.
We empirically set $\theta_c = 0.6$.

To evaluate the performances, the mean Average Precision~(mAP) and $F_1$ score of predicted object paths are measured.
The \texttt{Vehicle-3I} dataset includes 1,630 training IDs and 408 test IDs, consistent with the previous experiments in Section.~\ref{sec:exp}.
We evaluated our methods in real-world scenarios using the \texttt{Vehicle-3I} dataset.
For the evaluation, each test query was performed progressively using ReID according to the estimated connection matrix. 
The ReID process for each query continued until no identities satisfied the condition $\theta_c>0.6$ between the test query and its gallery.
In this way, the ReID of all test queries is performed, and the performances~(mAP, $F_1$) are measured.

\begin{figure}
	\centering
	\includegraphics[width = 0.7
	\linewidth]{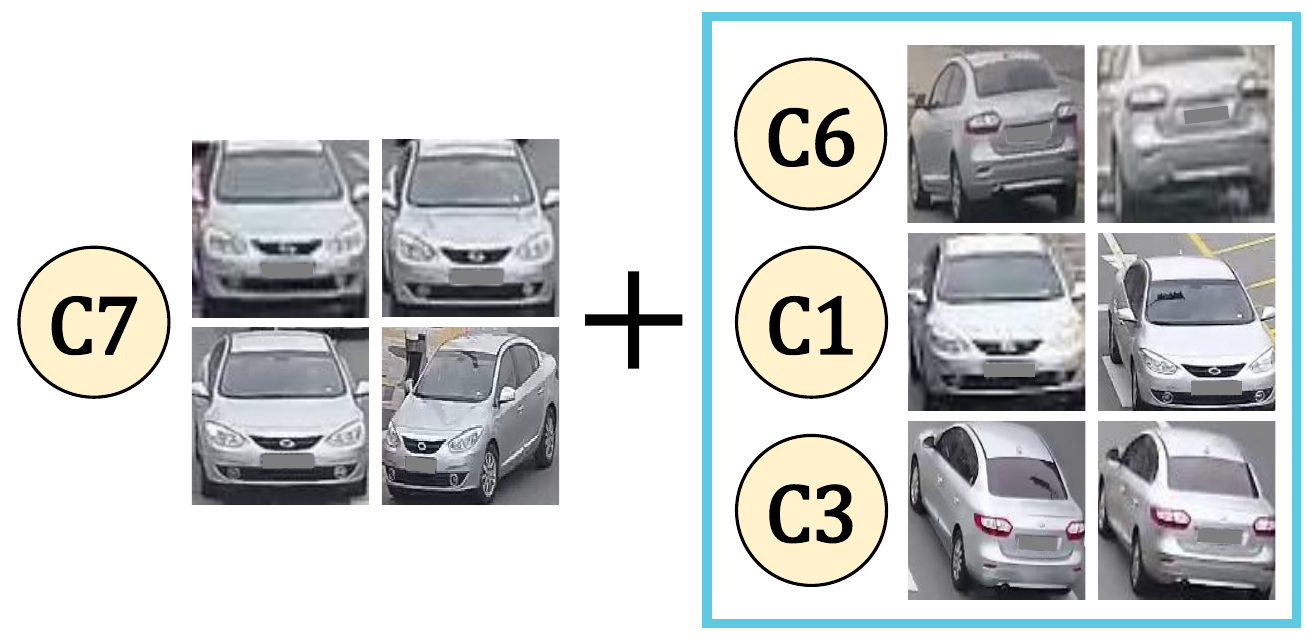}
	\caption{An example of the progressive query merging. The appearances in camera 7~(C7) are the initial query. The blue box denotes merged appearances in difference cameras~(C6,C1,C3) after performing Causal Identity Matching~(CIM).}
	\label{fig:query_set}
\end{figure}

\subsection{Effects of Causal Identity Matching~(CIM)}
The connection matrix for the \texttt{Vehicle-3I} dataset can be estimated using the paths of training identities. The size of the connection matrix $\mathbf{C}_{3I}$ for \texttt{Vehicle-3I} is $11\times11$.
This matrix is sparse, with only a few adjacent connections between cameras.
This means that when performing ReID for the target object, the entire distributions $p_{ij}\left(t\right)$ are not needed; only some adjacent $p_{ij}\left(t\right)$ are considered at each ReID step.
Furthermore, as shown in Fig.~\ref{fig:query_set}, the appearances of the target object become much richer through CIM and progressive query merging, allowing for more accurate ReID. 
Although a single camera can capture only fixed poses of the object, the query merging stage in CIM easily overcomes this limitation.
It increases the overall efficiency and accuracy of the ReID system, as in Table~\ref{tab:query_set_tab}.
This is a very natural procedure, but note that most recent studies have not closely examined such real-world conditions and causality for ReID.

\begin{table}[]
\centering
\renewcommand{\arraystretch}{0.95} 
\setlength\tabcolsep{3.8pt}
    \fontsize{8.2}{10}\selectfont
    \begin{tabular}{l||c|c}
    \noalign{\hrule height 1pt}
        Method & mAP & $F_1$ \\ \hline
        w/o Causal Identity Matching & 93.52 & 94.14 \\ \hline
        w/ Causal Identity Matching & 94.95 & 95.19 \\
    \noalign{\hrule height 1pt}
    \end{tabular}
    \caption {ReID performances without and with CIM.}
\label{tab:query_set_tab}
\end{table}

\subsection{Effects of Top$-k$ multi-shot matching}
In this experiment, we test the effect of Top$-k$ multi-shot matching between two objects. To this end, we tried different matching strategies as follows:
\begin{itemize}
	\item
	Single-shot: performing ReID that only uses a single appearance for each object. 
	For unbiased evaluations, we randomly selected a single appearance for each identity and repeated the appearance selection ten times to average them for the final score. 
	\item
	Multi-shot~(max): matching all possible pairs between multiple appearances and selecting the smallest matching score for the final score as used in \cite{farenzena2010person}.
	\item
	Multi-shot~(avg.): matching all possible pairs between multiple appearances and averaging all matching scores as used in~\cite{li2015multi}.
	\item
	Multi-shot~(Top-$k$): proposed method -- matching all possible pairs between multiple appearances and averaging only top-$k$ scores.
\end{itemize}
As summarized in Table.~\ref{tab:result_2}, we compared several top-$k$ multi-shot matching methods including $k=5,10,20$, and tested the ReID models FastReID~\cite{he2020fastreid} and ours. 

In both methods~(FastReID~\cite{he2020fastreid}, Ours), the single-shot matching strategy performed lowest accuracies due to the limited appearance cues. Moreover, using a single appearance for each identity does not correspond to real-world scenarios.
Compared to the `single-shot', multi-shot~(min) and multi-shot~(avg.) strategies achieved higher performances. 
This results support that using multiple appearances improves ReID performances.
However, multi-shot~(max) can be highly biased in the best similarity score between two sets of appearances.
Meanwhile, multi-shot~(avg.) can consider multiple comparisons, but the final score can be smoothed out or polluted by unreliable appearances. 

On the other hand, the proposed Top-$k$ matching is neither too biased nor too smoothed, and it produces a stable score even when the number of compared appearances changes by appearance management as proposed in Section~\ref{sec:proposed_4}.
Regardless of the $k$ value, the proposed Top-$k$ multi-shot matching methods achieved higher performance than other strategies such as single and multi-shot (max, avg.) in both methods~(FastReID~\cite{he2020fastreid}, Ours).
It achieved a 94.95\% mAP and a 95.19 \% $F_1$ score when $k=5$.

\subsection{Effects of appearance management}
The complexity of multi-shot matching between two appearance sets is $O\left(N_mN_n\right)$, where $N_m, N_n$ are the appearance numbers of $m$- and $n$-th objects. This roughly follows a complexity of $O\left(N^2\right)$ and needs to be improved for efficiency.
Additionally, many duplicated and noisy appearances degrade performance, as discussed in Section~\ref{sec:proposed_4}.
To address this, we utilized the proposed appearance unreliability score $\mathcal{U}\left(\mathbf{I}^m\right)$ to adjust the number of appearances for each object.

In Table~\ref{tab:result_3}, we tested different settings for the proposed appearance management and performed Top-$5$ multi-shot matching. The App. selection ratio means the proportion of collected appearances used for ReID. For example, we can use only a few reliable appearances such as $5-50\%$ according to the unreliability score $\mathcal{U}\left(\mathbf{I}^m\right)$.
App. selection ratio of `ALL' used all collected appearances without performing the proposed appearance management.
The proposed appearance management improved ReID performances by approximately 2\% and 1.32\% in mAP scores for the FasetReID and our method, respectively, when only $5\%$ of reliable appearances were utilized.
These results indicate that, regardless of the ReID methods used, ReID performance can be improved with only a few selected appearances.
In addition, the matching complexities are dramatically reduced. 
Additionally, matching complexities are significantly reduced. For example, when selecting only 5\% of appearances, 99.75\% of redundant calculations can be eliminated, while maintaining or even enhancing performance.

\begin{table}[]
\centering
\renewcommand{\arraystretch}{0.95} 
\setlength\tabcolsep{3.8pt}
    \fontsize{8.2}{10}\selectfont
    \begin{tabular}{c|l||c|c}
       \noalign{\hrule height 1pt}
       \multicolumn{2}{r||}{Datasets} & \multicolumn{2}{c}{\texttt{Vehicle-3I}} \\ \hline
        Models                                          & Matching methods    & mAP   & $F_1$ score  \\ \hline
        \multirow{6}{*}{FastReID~\cite{he2020fastreid}} & single-shot         & 79.71   & 80.08  \\ \cline{2-4}
                                                        & multi-shot~(max)    & 86.32    & 86.55 \\ \cline{2-4}
                                                        & multi-shot~(avg.)   & 87.14    & 87.45 \\ \cline{2-4}
                                                        & multi-shot~(Top-5)  & 87.46    & 87.73 \\ \cline{2-4}
                                                        & multi-shot~(Top-10) & 88.11    & 88.42 \\ \cline{2-4}
                                                        & multi-shot~(Top-20) & $\mathbf{88.55}$    & $\mathbf{88.90}$ \\ \hline
        \multirow{6}{*}{Ours}                           & single-shot         & 88.75    & 89.24 \\ \cline{2-4}
                                                        & multi-shot~(max)    & 94.30    & 94.57 \\ \cline{2-4}
                                                        & multi-shot~(avg.)   & 91.15    & 91.65 \\ \cline{2-4}
                                                        & multi-shot~(Top-5)  & $\mathbf{94.95}$    & $\mathbf{95.19}$ \\ \cline{2-4}
                                                        & multi-shot~(Top-10) & 94.17    & 94.48 \\ \cline{2-4}
                                                        & multi-shot~(Top-20) & 94.17    & 94.54 \\ 
    \noalign{\hrule height 1pt}
    \end{tabular}

    \caption {Performance comparisons of various matching strategies on the \texttt{Vehicle-3I} dataset. Single-shot and multi-shot~(max, avg.) are the traditional ReID matching strategies, and multi-shot (top-$k$) is our proposed matching method. The best performances are marked in \textbf{bold}.}
\label{tab:result_2}
\end{table}

\begin{table}[]
   \centering
   \renewcommand{\arraystretch}{0.95} 
   \fontsize{8.2}{10}\selectfont
   \begin{tabular}{c|c||c|c}
      \noalign{\hrule height 1pt}
      \multicolumn{2}{r||}{Datasets} & \multicolumn{2}{c}{\texttt{Vehicle-3I}} \\ \hline
      Models                                          & App. selection ratio  & mAP   & $F_1$ score \\ \hline
      \multirow{5}{*}{FastReID~\cite{he2020fastreid}} & ALL         & 85.47    & 85.72          \\ \cline{2-4}
                                          & 50 \%       & 84.80    & 85.02          \\ \cline{2-4}
                                          & 20 \%       & 86.57    & 86.81          \\ \cline{2-4}
                                          & 10 \%       & 87.22    & 87.49          \\ \cline{2-4}
                                          & 5  \%       & $\mathbf{87.46}$    & $\mathbf{87.73}$          \\ \hline
        \multirow{5}{*}{Ours}                           & ALL         & 93.63    & 93.85          \\ \cline{2-4}
                                          & 50 \%       & 91.91    & 92.24          \\ \cline{2-4}
                                          & 20 \%       & 93.95    & 94.24          \\ \cline{2-4}
                                          & 10 \%       & 94.77    & 94.93          \\ \cline{2-4}
                                          & 5  \%       & $\mathbf{94.95}$    & $\mathbf{95.19}$          \\ 
      \noalign{\hrule height 1pt}
   \end{tabular}
   
   \caption {ReID performances according to the App. selection ratio for the proposed appearance management. The best performances are marked in \textbf{bold}.}
   \label{tab:result_3}
\end{table}


\section{Conclusion}
This study proposes a ReID framework integrating a spatial-temporal fusion network with causal identity matching~(CIM) to address real-world challenges. 
The framework estimates camera network topology and combines appearance with spatial-temporal similarities to reduce appearance ambiguity. 
To this end, we proposed an adaptive Parzen window for reliable topology estimation and FusionNet for optimal similarity aggregation. 
The proposed framework achieves remarkable performance on the \texttt{VeRi776}~\cite{liu2016deep}, \texttt{Vehicle-3I} and \texttt{Market-1501}~\cite{zheng2015scalable} datasets, with rank-1 accuracies of 99.70\%, 89.22\%, and 99.11\% and mAP scores of 91.71\%, 83.62\%, and 95.5\% mAP, respectively.
The results indicate that incorporating spatial and temporal information in ReID can improve the accuracy of appearance-based methods and effectively address appearance ambiguity.

Additionally, we proposed causal identity matching~(CIM) to consider the issues in real-world scenarios.
The proposed CIM leverages adjacent cameras and transition time distributions to dynamically assign gallery sets for target objects to include only relevant objects and progressively merge queries to improve ReID accuracy and variety.
On the \texttt{Vehicle-3I} dataset, CIM achieved notable performances with a 94.95\% mAP and 95.19\% F1 score.
The results demonstrate that the proposed CIM is effective in real-world surveillance scenarios beyond the laboratory setting. 
Although the proposed ReID framework utilizes lightweight networks, it effectively performs ReID tasks.  
The proposed CIM is flexible and can be integrated with other ReID frameworks and appearance-based methods.

\bibliographystyle{IEEEtran}
\bibliography{main}

\vfill

\end{document}